\documentclass[journal]{IEEEtran}
\usepackage{graphicx}
\usepackage{amssymb}
\usepackage{amsmath}
\usepackage{multirow}
\usepackage{cite}
\usepackage{booktabs}
\usepackage{subfigure}
\usepackage{diagbox}
\usepackage{color}

\usepackage{xspace}
\makeatletter
\DeclareRobustCommand\onedot{\futurelet\@let@token\@onedot}
\def\@onedot{\ifx\@let@token.\else.\null\fi\xspace}

\def\etal{\emph{et al}\onedot}
\makeatother

\ifCLASSINFOpdf
\else
\fi
\begin{document}
%
\title{Dizygotic Conditional Variational AutoEncoder for Multi-Modal and Partial Modality Absent Few-Shot Learning}
%
%
%

\author{Yi Zhang,
  Sheng Huang,~\IEEEmembership{Member,~IEEE,}
  Xi Peng,~\IEEEmembership{Member,~IEEE,} and~Dan~Yang
  \thanks{This work was supported in part by the National Key Research and Development Project under Grant 2018YFB2101200, in part by the National Natural Science Foundation of China under Grant 61772093, and in part by the Science and Technology Research Program of Chongqing Municipal Education Commission of China under Grant KJQN201900726. (S. Huang is the corresponding author.)}
  \thanks{S. Huang is with Ministry of Education Key Laboratory of Dependable Service Computing in Cyber Physical Society, Chongqing, 400044, P.R.China,
    Y. Zhang, S. Huang, D. Yang are with the School of Big Data and Software Engineering, Chongqing University, Chongqing, 400044 P.R.China,(email:\{zhangyii, huangsheng, dyang\}@cqu.edu.cn),}
  \thanks{X. Peng is with Department of Computer and Information Sciences at University of Delaware, Newark, DE 19716, USA, (email:xipeng@udel.edu)}
}

\maketitle

\begin{abstract}
  Data augmentation is a powerful technique for improving the performance of the few-shot classification task. It generates more samples as supplements, and then this task can be transformed into a common supervised learning issue for solution. However, most mainstream data augmentation based approaches only consider the single modality information, which leads to the low diversity and quality of generated features. In this paper, we present a novel multi-modal data augmentation approach named Dizygotic Conditional Variational AutoEncoder (DCVAE) for addressing the aforementioned issue. DCVAE conducts feature synthesis via pairing two Conditional Variational AutoEncoders (CVAEs) with the same seed but different modality conditions in a dizygotic symbiosis manner. Subsequently, the generated features of two CVAEs are adaptively combined to yield the final feature, which can be converted back into its paired conditions while ensuring these conditions are consistent with the original conditions not only in representation but also in function. DCVAE essentially provides a new idea of data augmentation in various multi-modal scenarios by exploiting the complement of different modality prior information. Extensive experimental results demonstrate our work achieves state-of-the-art performances on miniImageNet, CIFAR-FS and CUB datasets, and is able to work well in the partial modality absence case.
\end{abstract}


%
\IEEEpeerreviewmaketitle

\section{Introduction}
\IEEEPARstart{I}{n} recent years, with the upsurge of research in the field of artificial intelligence, the supervised learning technique achieves extensive remarkable successes due to the significant advance of deep learning and the availability of large scale labeled data. However, some studies indicate that the frequencies of observing objects often follow a long-tailed distribution in the wild, and the number of rare objects significantly surpasses that of common objects~\cite{longtail}. Moreover, labeling data is costly in labor and time, since some tasks may need fruitful expert knowledge for data labeling, such as fine-grained species recognition and malignancy assessment. In other words, the supervised learning task is more realistic in few-shot scenarios where some categories do not have enough labeled data for training. This fact leads to the emergence of few-shot learning technique which understands the new concept with only limited relevant labeled examples and a large number of irrelevant labeled data.

\begin{figure}[tbp]
  \centering
  \includegraphics[scale=0.65]{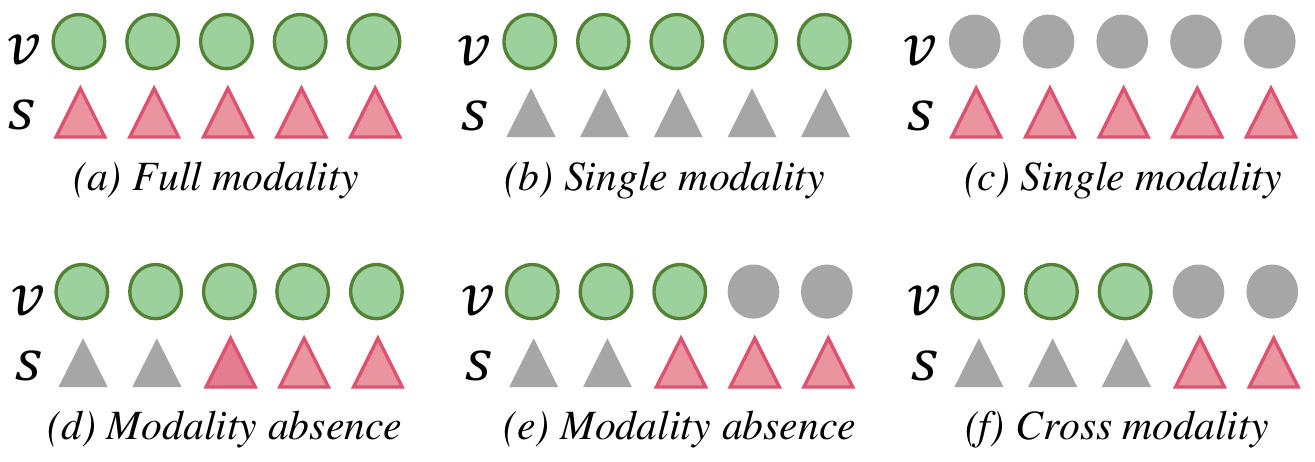}
  \vspace{-0.2cm}
  \caption{Several data modality configurations in the few-shot image classification stage. ``$v$'' and ``$s$'' represent visual and semantic modality information, respectively. (a) Training with full modality~\cite{xing2019adaptive,pahde2021multimodal}; (b-c) Training with single modality~\cite{snell2017prototypical,xian2019f}; (d-f) We additionally study more challenging configurations of modality absence and cross modality cases. Green and pink indicate the existence of modality data, while gray indicates the absence of the corresponding modality data. (Best viewer in color.)}
  \label{modal}
  \vspace{-0.5cm}
\end{figure}

Over the past few years, an increasing number of few-shot learning approaches have been proposed to address this challenge. These methods are mainly divided into three categories. The first one is the metric based approach~\cite{vinyals2016matching,snell2017prototypical,khrulkov2020hyperbolic}, which aims to learn a metric space for pulling the homogenous samples closer to each other while pushing the inhomogeneous samples away. Another category is the gradient based method~\cite{finn2017model,rusu2018meta,simon2020adaptive}. The basic idea of this method is learning to learn which attempts to train a meta-learner with a good generalization ability based on a large number of tasks, and quickly adapt the model to the new task with only a few labeled samples and a small number of gradient update steps. The last one is the data augmentation based method~\cite{zhang2018metagan,zhang2019few,li2020adversarial,chen2020diversity}, which attempts to generate the extra samples for the few-shot categories to mitigate the lack of the labeled samples.

Data augmentation is deemed as the panacea for alleviating data scarcity, and it has become a novel way for few-shot learning very recently. The early works focus on performing a variety of deformations on the image~\cite{Zitian,chen2019image} to produce the extra samples. However, the generated samples based on this scheme lack diversity and no prior information is implanted into the sample generation procedure, which leads to its low performance in comparison with the other kinds of few-shot learning approaches. Another kind of data augmentation method is to employ the conditioned deep generative model for sample generation which gains more attention recently in few-shot learning~\cite{zhang2018metagan,li2020adversarial,chen2020diversity}. The main drawback of these methods is that they only consider the single modality information, mostly visual information, for conditioning the feature generation, as the data settings illustrated in Figure \ref{modal}(b). However, in the few-shot learning,  particularly the one-shot scenario, it is pretty hard to leverage the limited visual information only from a single sample to guide the generation of varying features.

Inspired by the recent advance of Natural Language Processing (NLP), the semantics, as an easily-accessible information, plays an increasingly important role in visual learning tasks. It is frequently used for manipulating the feature generation in zero-shot learning~\cite{xian2017zero,xian2019f,avae}, which essentially follows the data settings illustrated in Figure~\ref{modal}(c), and achieves the impressive performance via considering it as the complementarity of visual information in many few-shot learning approaches~\cite{xing2019adaptive,schwartz2019baby}. These approaches can be essentially deemed as the multi-modal few-shot learning approach under the data settings as illustrated in Figure~\ref{modal}(a). For the case of few-shot learning under some more general or extreme data settings, such as modality absence or cross-modal scenarios shown in Figure~\ref{modal}(d)-(f), this issue still remains unstudied. However, these scenarios are actually quite frequent occurrence in many uncontrolled environments~\cite{tsai2018learning,shi2021relating,ma2021smil}.

To address the aforementioned issues, we present a novel multi-modal deep generative learning framework named Dizygotic Conditional Variational AutoEncoder (DCVAE) which focuses on fully utilizing both the semantic and visual information as multi-modal conditions to generate the features for few-shot learning. DCVAE pairs two Conditional Variational AutoEncoders (CVAEs) to generate a pair of features with the same seed but the different modality conditions, and then a dizygotic feature adaptive mixture module is employed to unify these features as a final synthetic feature in a linear manner. Meanwhile, such synthetic feature should be able to be transformed back to its paired conditions, and these retrieved conditions should be consistent with the original ones not only in representation but also in function. Three popular few-shot learning benchmarks, namely miniImageNet, CIFAR-FS and CUB, are used to evaluate our approach. Extensive results demonstrate DCVAE not only achieves the state-of-the-art performances, but also is able to work well under all data settings illustrated in Figure~\ref{modal}.

In summary, the main contributions of our work are as follows:
\begin{itemize}
  \item We propose an ingenious feature generation approach named DCVAE, which sufficiently utilizes multi-modal information with both the semantic and the visual to generate the high-quality features for few-shot learning. Extensive experimental results on three popular benchmarks and three well-known backbones show that our method achieves very promising performance in comparison with the recent state-of-the-art approaches.
  \item DCVAE provides a novel conditional VAE framework for data synthesis with the paired conditions. Such framework is also flexible to be further extended to support the multi-condition case (more than two conditions) via following the dizygotic symbiosis manner.
  \item DCVAE adopts an adaptive linear combination fashion to construct the final feature with different modality features with respect to each sample, which endows it with the ability to tackle the few-shot learning task under modality absence and cross-modal scenarios as illustrated in Figure~\ref{modal}, since the final feature and different modality features are all lying in the same space and available for comparing with each other.
\end{itemize}

\begin{figure*}[htp]
  \centering
  \includegraphics[scale=0.63]{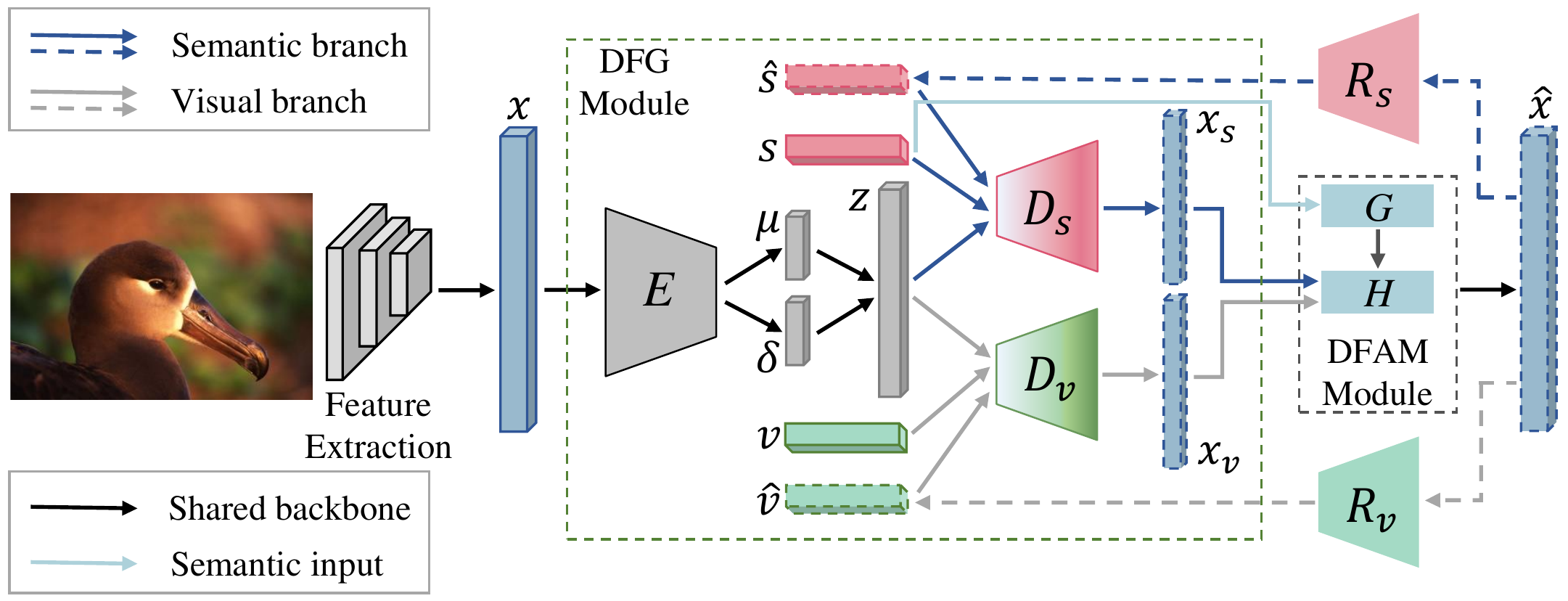}
  \vspace{-0.1cm}
  \caption{The framework of our proposed DCVAE model, which contains four modules: {\em Feature Extraction, Dizygotic Feature Generation (DFG), Dizygotic Feature Adaptive Mixture (DFAM) and Condition Cyclic Consistency (CCC)}. The {\em DFG module} pairs two CVAEs to generate a pair of features ($x_{s}$ and $x_{v}$) with the same seed but the different modality conditions (semantic embedding $s$ and class prototype $v$). The {\em DFAM module} adaptively unifies these two features as a final synthetic feature $\hat{x}$ in a linear manner. The {\em CCC module} ($R_{s}$ and $R_{v}$) aims to convert the fused feature $\hat{x}$ back into its paired conditions ($\hat{s}$ and $\hat{v}$) while ensuring these conditions are consistent with the original conditions not only in representation but also in function. (Best viewer in color.)}
  \label{fig1}
  \vspace{-0.4cm}
\end{figure*}

\vspace{-0.3cm}
\section{Related Work}
In this section, we review the mainstream methods of few-shot learning, which can be grouped into {\em metric based} methods, {\em gradient based} methods and {\em data augmentation based} methods. In addition, we also provide a brief review of few-shot learning approaches, which involve multiple modality information.

\vspace{-0.3cm}
\subsection{Metric Learning Based Methods}
Metric learning is a classic way for few-shot learning, which aims to learn a metric space for distinguishing samples with their similarities. The early work of metric learning focused on measuring the distance from each query sample to its support images through some simple metrics, such as cosine distance or Euclidean distance~\cite{vinyals2016matching,qi2018low,tadam}. Instead of using these simple metrics, Relation Network~\cite{sung2018learning} leverages a neural network for learning the similarity between the query and its support samples. Different to the aforementioned methods that measure similarities between samples, Prototypical Network~\cite{snell2017prototypical} measures the distance between the query sample and each class prototype for classification.
More recently, SMAVI~\cite{abdelaziz2021few} utilizes saliency maps as extra visual information and then combines image features and saliency map features into new features for classification.
Although these approaches have achieved encouraging performance, most of them are limited by the choice of metric function and embedding spaces, which leads to the poor generalization ability of the model.

\vspace{-0.3cm}
\subsection{Gradient Based Methods}
Gradient based methods aim to learn a model for learning the meta knowledge of tasks which helps the model to quickly adapt to recognize novel classes with limited samples. The gradient based methods have an intersection with the aforementioned metric based methods which adopt the specifically designed training strategy for learning the meta knowledge cross-tasks~\cite{vinyals2016matching,snell2017prototypical,sung2018learning}. Similarly, some approaches attempt to acquire prior knowledge by building meta-learner~\cite{rusu2018meta,ren2019incremental,sun2019meta} via a diverse set of tasks, and then use this knowledge to evaluate the final classification task. Another branch of this method is the optimization-based meta-learning approach~\cite{finn2017model,simon2020adaptive,oh2021boil}. It aims to learn a model which can quickly adapt to new tasks just with a small amount of fine-tuning and a suitable initialization. The main drawback of these methods is that they must sample a mass of tasks to learn the suitable meta-knowledge transfer, which is quite time-consuming.

\vspace{-0.3cm}
\subsection{Data Augmentation Based Methods}
Data augmentation is a more intuitive and straightforward approach, which transforms the few-shot classification into a common supervised learning problem by generating more samples for complementary. The early works generate the extra samples via simply applying a set of transformations (such as rotation, deformation, Gaussian perturbation and so on) to the original image~\cite{chen2019image,wang2018low,Zitian}. However, these methods are hard to generate the diverse samples and no prior knowledge has been incorporated into the sample generation procedure.

Recently, some more sophisticated data augmentation methods have been developed for few-shot learning and achieve more superior performance. For example, Li \etal~\cite{li2020adversarial} employs the conditional GAN to synthesize the data with prior visual information. Zhang \etal~\cite{zhang2019few} utilizes a saliency network for extracting the foregrounds and backgrounds of available images and feeds the resulting maps into a two-stream network to hallucinate data points directly in the feature space from viable foreground-background combinations. VI-Net~\cite{luo2021few} construct a generative framework to define a generating space for each class in the latent space based on support samples and then generating synthetic features based on it. Diversity Transfer Network~\cite{chen2020diversity} is presented to learn to transfer latent diversities from known categories and composite them with support features to generate diverse samples for novel categories. However, only the single modality prior knowledge (condition), mostly the visual information, is considered as the prior knowledge for manipulating the data generation in such methods. Unfortunately, it is unrealistic to extract the rich prior knowledge only from a few samples or even one sample for generating the diverse features.

\vspace{-0.3cm}
\subsection{Multi-Modal Few-Shot Learning}
The universality of modalities and the rapid development of deep learning endow multi-modal learning with great potential. Especially in the case of extremely scarce samples, multi-modal data can provide more information to the model to obtain the more comprehensive feature with a better generalization ability.

Recently, some works have leveraged multi-modal information to improve the performance of models in the context of few-shot learning. Dual TriNet~\cite{chen2019multi} maps the sample to the semantic space and adds noise for perturbation, and then maps the perturbation semantic samples back to the visual space to generate sample features. Xing \etal~\cite{xing2019adaptive} and Schwartz \etal~\cite{schwartz2019baby} both adopt an adaptive modality mixing mechanism, which further enhanced the existing classification algorithm based on metric learning by adaptively combining information from both semantic and visual modalities. Xian \etal~\cite{xian2019f} propose a conditional generation model which learns the marginal feature distribution of unlabeled images using an unconditional discriminator. Based on Prototypical Network~\cite{snell2017prototypical}, Pahde \etal~\cite{pahde2021multimodal} train a generative model to map semantic information into the visual feature space to obtain more reliable prototypes. These methods achieve very competitive performances with the help of multi-modal complementary information. However, all the methods only consider the ideal multi-modal data scenario where all the modalities of information are available for each sample. In the real world, the few-shot learning model often faces some more complex and challenging multi-modal data settings. For example, some modalities of samples may be unavailable or missing as illustrated in Figure~\ref{modal}(d), (e). In some more extreme cases, the modalities of samples from different categories may be even not overlapped with each other during testing as illustrated in Figure~\ref{modal}(f).

Although there have been a few of research efforts in handling missing modalities for multi-modal learning, such as testing with missing modality~\cite{tsai2018learning}, training with unpaired modality~\cite{shi2021relating}, or their mixture~\cite{ma2021smil}, the multi-modal few-shot learning in the above extreme cases still remains unstudied, which severely limits its effectiveness in some more realistic scenarios.

\vspace{-0.1cm}
\section{Methodology}
\vspace{-0.1cm}
\subsection{Problem Definition and Notations}
In the few-shot setting, there is no intersection between the class spaces of the source domain and the target domain, and the target domain only contains a few labeled data. Therefore, the few-shot learning task is to learn the concept of novel categories with a small amount of labeled samples in the target domain based on the source domain which has sufficient labeled samples from non-intersect categories.

Let $\mathcal{D}_{train} = \{(x,y,s_{y})|x\in X^{tr},y\in Y^{tr},s_{y}\in A^{tr}\}$  and  $\mathcal{D}_{test} = \{(x,y,s_{y})|x\in X^{te},y\in Y^{te},s_{y}\in A^{te}\}$ be the training and test sets in the source/target domain respectively, where their class sets are disjoint.
$x$ is denoted as a visual feature produced by feature extractor and $y$ is its corresponding label. $s_{y}$ is the semantic representation for class corresponding to the label $y$. $X^{tr}$, $Y^{tr}$ and $A^{tr}$ are denoted as the feature, label and semantic embedding sets of training data respectively. According to the definitions in literatures~\cite{vinyals2016matching,snell2017prototypical}, a few-shot classification task is often called $N$-way $K$-shot problem. Each $N$-way $K$-shot classification task $\mathcal{T}$ consists of a support set $\mathcal{S_{\mathcal{T}}}$, which contains $K$ labeled samples for each of the $N$ classes, and a query set $\mathcal{Q_{\mathcal{T}}}$ where the samples are randomly sampled from the same $N$ classes on the test dataset. Thus, we can define the task as: $\mathcal{T} = \{(\mathcal{S_{\mathcal{T}}},\mathcal{Q_{\mathcal{T}}})\}$ where $\mathcal{S_{\mathcal{T}}} = \{(x_{i},y_{i},s_{y_{i}})|x_{i}\in X^{te},y_{i}\in Y^{te},s_{y_{i}}\in A^{te}\}_{i=1}^{N\times K}$ and $\mathcal{Q_{\mathcal{T}}} = \{(x_{j},y_{j})|x_{j}\in X^{te},y_{j}\in Y^{te}\}_{j=1}^{M}$. Here, $M$ is the number of samples in the query set. $X^{te}$, $Y^{te}$ and $A^{te}$ respectively denotes the feature, label and semantic embedding sets of test data.

In the modality absence scenarios, some modalities of a sample are partially missing, e.g., $(x_i,y_{i},-)$ or $(-,y_{i},s_{y_i})$. The cross-modal scenario is a special case of the modality absence as seen in Figure~\ref{modal}(f), e.g., $\{(x_i,y_i,-)\}\cap\{(-,y_j,s_{y_j})\} = \varnothing$. Here, we consider that these modality absence cases only exist in the support set $\mathcal{S_{\mathcal{T}}}$, since the categories of training data have no overlap with the ones of testing data, and the training data can be manually collected by users, which are able to collect the samples from categories with full modalities only.

\vspace{-0.3cm}
\subsection{Overview}
We present a novel deep generative approach named \textbf{D}izygotic \textbf{C}onditional \textbf{V}ariational \textbf{A}uto\textbf{E}ncoder (\textbf{DCVAE}) for few-shot learning from the perspective of data augmentation. The framework architecture of our method is illustrated in Figure \ref{fig1}, which consists of four components: Feature Extraction, Dizygotic Feature Generation (DFG), Dizygotic Feature Adaptive Mixture (DFAM) and Condition Cyclic Consistency (CCC). Firstly, only the training data is used to train the feature extraction network to extract discriminative image features. The key component of our model is the dizygotic feature generation. It is a paired conditional variational autoencoder model which leverages an encoder $E(\cdot)$ to convert the input visual feature $x$ into a latent representation encoded by the gaussian distribution parameters $\mu$ and $\sigma$, and then uses two decoders $D_{s}(\cdot)$ and $D_{v}(\cdot)$ to reconstruct the input feature $x$ respectively with different conditions and the random noise vector $z$ generated by $\mu$ and $\sigma$. After that, we can procure the final synthetic feature $\hat{x}$ via fusing the pair of the reconstructed features generated by decoders in an adaptive feature mixture mechanism~\cite{xing2019adaptive}. In pace with the generator, a condition cyclic consistency module is introduced for transforming the final synthetic feature $\hat{x}$ back to its corresponding conditions. Moreover, the dizygotic feature generation procedure should be conducted again with these reconstructed conditions to validate the condition cyclic consistency.

\vspace{-0.3cm}
\subsection{Feature Extraction}
The feature extraction is the basis of feature generation algorithms to success. The Convolutional Neural Networks (CNNs) are considered as one of the most dominant feature learning methods which have been widely used in few-shot learning. The mainstream FSL methods~\cite{xing2019adaptive,avae,yoon2019tapnet,chen2020diversity,li2020adversarial} mainly adopt ResNet with different depths (such as ResNet-12, ResNet-18 and ResNet-101, etc.) to extract features. Here, we choose ResNet-12, a relatively lightweight network, as the basic feature extractor of our framework.

Meanwhile, ResNet-18 is also frequently adopted in few-shot learning, so we use it as another feature extraction network for miniImageNet and CIFAR-FS. With regard to CUB, the dataset provides off-the-shelf deep features extracted by ResNet-101~\cite{xian2017zero} that have been adopted by some few-shot learning methods~\cite{xian2019f,avae,xing2019adaptive}, therefore we also use these features for verification on CUB datasets.

\vspace{-0.3cm}
\subsection{Dizygotic Feature Generation}
In our paper, we intend to generate the deep features for the few-shot categories and then consider few-shot learning task as an ordinary classification issue for solution. Variational AutoEncoder (VAE) and Generative Adversarial Network (GAN) are two of the most popular generative frameworks. Here, we choose VAE as the basic generative network, since some works indicated that the samples generated by GAN suffer from the mode collapse and lack of diversity~\cite{razavi2019generating}.

The conventional CVAE contains an encoder $P_{E}(z|x,c)$ that maps the feature $x$ into the latent representation conditioned by class embedding $c$, and a decoder $P_{D}(x|z,c)$ that reconstructs the input $x$ from the latent $z$ and condition $c$. However, this model is hard to be trained due to the fact that CVAE suffers from the posterior collapse problem~\cite{zhao2018unsupervised,makhzani2015adversarial}. In order to alleviate this problem, we follow the setting in \cite{verma2018generalized} to generate latent representation $z$ only by visual feature $x$. Thus, the objective function of CVAE can be formulated as,
\begin{equation}
  \begin{split}
    \mathcal{L_{\mathsf{c} \mathsf{v} \mathsf{a} \mathsf{e} }}=&~\mathbb{E}_{P_{data}(x,z),P_{E}(z|x)}[\log P_{D}(x|z,c)] \\
    &-\lambda \cdot D_{KL}(P_{E}(z|x)\| P(z)),
  \end{split}
\end{equation}
where $D_{KL}(\cdot \|
  \cdot)$ is the Kullback-Leibler divergence (KL-Divergence), $\lambda$ is a positive hyper-parameter for reconciling the reconstruction error and KL-Divergence, and $P(z)$ follows the distribution of $\mathcal{N}(0,1)$.

As shown in the green dashed box in Figure \ref{fig1}, our model generates discriminative and high-quality synthetic features conditioned by different prior knowledge, such as semantic embedding and class prototype, associated with the same random noise into the decoders respectively. {\em Such feature generation procedure is just like the procedure that two eggs are fertilized by the same seed and grow as twins that we name it Dizygotic Symbiosis Manner (DSM).} In such a manner, our feature generation network which is based on Conditional Variational AutoEncoder (CVAE) is named as Dizygotic Conditional Variational AutoEncoder (DCVAE).

The existing VAE or GAN-based few-shot and zero-shot learning approaches all only consider the single modality information as the condition for feature generation~\cite{xian2019f,li2020adversarial}. However, as two most important cues for feature generation, no matter the visual information or the semantic information have been proven its effectiveness in few-shot and zero-shot learning. Moreover, in the one-shot scenarios, it is almost impossible to synthesize the feature for a class with high diversity only based on the limited visual information from one sample. DCVAE aims to incorporate all these two cues as the twin conditions for generating the high-quality features.

DCVAE utilizes a shared encoder $E(\cdot)$ to learn a latent representation space parameterized by $\mu$ and $\sigma$, and then employs these parameters to randomly generate a latent representation $z$ as the random noise vector. Two decoders are leveraged for generating two independent features via respectively feeding the semantic information $s$ and visual information $v$ together with the same random noise vector $z$ where the visual-based decoder $D_v(\cdot)$ generates the visual-based synthetic feature $x_v$ while the semantic-based decoder $D_s(\cdot)$ generates the semantic-based synthetic feature $x_s$. $x_s$ and $x_v$ are the twins generated by the same seed but with different conditions. Here, the semantic representation of class such as attribute or word embedding is considered as the semantic information while the class prototype is deemed as the visual information. Following the work~\cite{snell2017prototypical}, the class prototype is directly generated by averaging the features from the same class:
\begin{equation}
  \begin{split}
    v_{k} = \frac{1}{\left\lvert C_{k}\right\rvert } \sum_{(x_{i},y_{i})\in C_{k} }x_{i},
  \end{split}
\end{equation}
where $C_{k}$ denotes the set of examples labeled with class $k$. In such a way, the loss of CVAE should be further updated as follows in our case,
\begin{equation}
  \begin{split}
    \mathcal{L_{\mathsf{b} \mathsf{c} \mathsf{v} \mathsf{a} \mathsf{e} }}=&~\mathbb{E}_{P_{data}(x,z),P_{E}(z|x)}[\log P_{D_{s}}(x|z,s)] \\
    +&~\mathbb{E}_{P_{data}(x,z),P_{E}(z|x)}[\log P_{D_{v}}(x|z,v)] \\
    -&~\lambda \cdot D_{KL}(P_{E}(z|x)\| P(z)), \label{bcvae}
  \end{split}
\end{equation}
where $\lambda$ is a positive hyper-parameter for weighting the KL-Divergence.

In addition, the similarity of the generated twin features should be maximized (the distance between them should be minimized), since they are all generated from the same seed (the random noise) and represent the same example. In such a manner, an additional twin similarity loss, which measures the Euclidean distance between the twin features, is introduced to DCVAE,
\begin{equation}
  \begin{split}
    \mathcal{L}_{ts} =\left\lVert D_{s}(s,z) - D_{v}(v,z) \right\rVert_{2}^{2}. \label{ts}
  \end{split}
\end{equation}

\vspace{-0.3cm}
\subsection{Dizygotic Feature Adaptive Mixture}
After obtaining the paired features $x_v$ and $x_s$, we need to unify them as a final unique feature $\hat{x}$. We adopt an adaptive mixture mechanism~\cite{xing2019adaptive} to accomplish this task which considers the final feature as an adaptive linear combination of features,
\begin{equation}
  \hat{x} = H(x_s,x_v)=\eta\cdot x_{s} + (1 - \eta)\cdot x_{v},
\end{equation}
where $H(\cdot)$ is such adaptive linear combination operation and $\eta$ is the adaptive mixture coefficient which is learned by a feature mixing network $G(\cdot)$ via giving the corresponding semantics $s$,
\begin{equation}
  \eta = \sigma (G(s)),
\end{equation}
where $\sigma(\cdot)$ is the sigmoid function.

  {\em This adaptive linear combination approach implicitly forces the features generated by the conditions of different modalities, i.e. $x_v$, $x_s$ and $\hat{x}$, to lie in the same feature space, and provides an intuitive way for comparing these aforementioned features with each other. In other words, DCVAE is capable of accomplishing the few-shot learning task even in the case that some modalities of examples are absence as illustrated in Figure~\ref{modal}.} Moreover, if only the visual information $v$ is available while the semantic information $s$ is absent, DCVAE is degenerated as the class prototype-conditioned VAE (the visual version). Similarly, if only the semantic information $s$ is available, DCVAE is degenerated as the semantic conditioned VAE, and the few-shot learning task is degenerated as the zero-shot learning task. With regard to the case that only some visual or semantic modality knowledge is missing, the original few-shot learning task is translated as the hybrid task of few and zero-shot learning.

\vspace{-0.3cm}
\subsection{Condition Cyclic Consistency}
In order to ensure that the final synthetic feature $\hat{x}$ generated by cross-modal conditions can fully encode both semantic and visual cues, we require the final synthetic feature $\hat{x}$ can be converted back to its corresponding original conditions $s$ and $v$ via the semantic consistency network $R_s(\cdot)$ and the visual consistency network $R_v(\cdot)$ respectively. Moreover, these retrieved conditions should be consistent with the original conditions not only in representation but also in function.

The condition consistency in representation means that the retrieved conditions should be as similar as possible to the original conditions. Meanwhile, as two of the most commonly used metrics in NLP and computer vision, cosine distance is usually used to measure the similarity between semantic features, while Euclidean distance is used to measure the similarity between visual features. Thus, we attempt to maximize the similarities between original conditions and retrieved ones via maximizing the cosine distances between semantic representations while minimizing the Euclidean distances between class prototypes. Finally, such a goal can be achieved by minimizing the following representation consistency loss,
\begin{equation}
  \mathcal{L}_{rc} =\frac{\left\lVert v - \hat{v} \right\rVert_{2}^{2}}{\cos (s, \hat{s}) + \epsilon}, \label{rc}
\end{equation}
where $\hat{s}=R_{s}(\hat{x})$ and $\hat{v}=R_{v}(\hat{x})$ are the retrieved semantic and visual conditions respectively. $\epsilon$ is a constant to avoid the denominator being divided by zero. Here, $\epsilon=0.1$.

The condition consistency in function means that the retrieved conditions should have the same effect as their original ones in feature generation. In other words, the final synthetic feature generated by the original semantic representation $s$ and the retrieved class prototype $\hat{v}$ should be consistent with the one generated by the retrieved semantic representation $\hat{s}$ and the original class prototype $v$. Such property can be achieved by minimizing the following function consistency loss,
\begin{equation}
  \mathcal{L}_{gfc} =\lVert H(\hat{x}_{s},x_{v}) - H(x_{s},\hat{x}_{v})\rVert_{2}^{2}, \label{gfc}
\end{equation}
where $\hat{x}_{s}=D_{s}(\hat{s},z)$ and $\hat{x}_{v}=D_{v}(\hat{v},z)$ are synthetic features generated by combining different retrieved condition and original condition respectively.

\vspace{-0.3cm}
\subsection {The DCVAE Model}
Integrating all losses in Equations~\ref{bcvae}, \ref{ts}, \ref{rc} and \ref{gfc} to yield the final DCVAE Model,
\begin{equation}
  \arg\underset{_{E,D,R,G}}\min~\mathcal{L_{\mathsf{D} \mathsf{C} \mathsf{V} \mathsf{A} \mathsf{E} }} =
  \mathcal{L_{\mathsf{b} \mathsf{c} \mathsf{v} \mathsf{a} \mathsf{e} }} + \mathcal{L}_{ts} + \mathcal{L}_{rc} + \mathcal{L}_{gfc}, \label{final}
\end{equation}
to achieve all the aforementioned feature generation properties.

\subsection{Details of Network Architecture}
Following the previous data augmentation works~\cite{xian2019f,li2020adversarial,chen2020diversity}, we adopt ResNet with different depths (such as ResNet-12, ResNet-18 and ResNet-101, etc.) as the backbone to extract features from all three datasets. This is done for the sake of fair comparison with the backbone of different data augmentation approaches.

The encoder $E(\cdot)$ of DCVAE consists of two fully connected layers of 1200 and 600 dimensions respectively, all followed by ReLU as the activation. Decoders $D_{s}(\cdot)$ and $D_{v}(\cdot)$ are implemented with one hidden layer of 600 units. In addition, each of the two branches of the cyclic consistency module is also implemented by one hidden layer of 512 units. The feature mixing network $G(\cdot)$ has only one hidden layer of 1024 units and the noise vector $z$ is a 100-dimensional vector drawn from the uniform Gaussian distribution.

During the feature extraction, the SGD optimizer is exploited with a momentum of 0.9 and a weight decay of $5e^{-4}$. The learning rate is initialized as 0.05 and is decayed after epoch 60 and 80 by a factor of 0.1, respectively. We choose Adam as our feature generate optimizer, where the learning rate is fixed to $10^{-4}$.

\vspace{-0.3cm}
\subsection{Training Strategies of DCVAE}
The training stage of few-shot learning model is commonly divided into two steps. The first step is to pre-train the few-shot learning model with training dataset. The second step is to fine tune the model with the support set and then infer label for the query set.

In the multi-modal few-shot learning, the modality absence case is more frequently happening in the fine-tuning step, since the categories of training and testing set can be non-overlapped, and thus we can especially collect samples with full modalities only to construct the training dataset in real scenarios. In such a manner, the pre-training step of DCVAE is as same as the one of other few-shot learning models. However, the fine-tuning step needs to face two situations in multi-modal case. One is the full modality case, the other is the modality absence case. In the full modality case, the fine-tuning step of DCVAE is as same as the one of the other approaches.

With regard to the fine-tuning step of DCVAE in the modality absence case, the batch-based optimization strategy should be switched to the subbatch-based optimization strategy, since the modalities of samples in the same batch may be different as shown in Figure~\ref{modal}(b)-(f). The batch is further divided into several subbatches based on the modality situations of samples, e.g., the full modality subbatch $\{x_i,y_i,s_{y_i}\}$, the visual modality absence subbatch $\{-,y_i,s_{y_i}\}$ and the semantic modality absence subbatch $\{x_i,y_i,-\}$.

For the full modality subbatch $\{x_i,y_i,s_{y_i}\}$, the parameters of all modules are updated during this subbatch-based optimization. For the semantic modality absence subbatch $\{x_i,y_i,-\}$, only the parameters of the encoder $E(\cdot)$ and the visual decoder $D_v(\cdot)$ are updated while other modules are frozen during this subbatch-based optimization.
However, due to the lack of visual samples, the semantic decoder $D_s(\cdot)$ is directly adopted to generate synthetic features for the visual modality absence subbatch $\{-,y_i,s_{y_i}\}$ setting.
The fine-tuning of DCVAE in the modality absence cases shown in Figure~\ref{modal} all can be accomplished by the different mixtures of these aforementioned subbatch-based optimizations.

\vspace{-0.3cm}
\subsection{Label Inference}\label{id}
Once the model has been fine-tuned, as illustrated in Figure \ref{fig1}, we can generate three kinds of synthetic features for each class in $\mathcal{S_{\mathcal{T}}}$, and then combine those synthetic features with the support set $\mathcal{S_{\mathcal{T}}}$. After that, any classifier can be trained based on this augmented support set. In this paper, we just apply the simple $k$-nearest neighbor($k$-NN) classifier to classify samples from the query set $\mathcal{Q_{\mathcal{T}}}$.

\begin{table*}[htbp]
  \caption{Few-shot classification accuracy(in \%) on CUB dataset. ``Others'' include metric and gradient based methods, ``DataAug'' represents data augmentation based methods. $^{\dagger }$ indicates that the method is a multi-modal approach, which has leveraged the additional semantic information. The best results are displayed in \textbf{boldface.}}
  \centering
  \small
  \setlength{\tabcolsep}{3mm}{
    \begin{tabular}{lccccc}
      \toprule
      Source       & Methods                                             & Type                     & Feature Extractors & 5-way 1-shot   & 5-way 5-shot   \\
      \midrule
      ICCV'2019    & SAML~\cite{hao2019collect}                          & \multirow{11}{*}{Others} & 64-64-64-64        & 69.35          & 81.37          \\
      ICLR'2019    & Closer Look~\cite{chen2019closer}                   &                          & 64-64-64-64        & 60.53          & 79.34          \\
      NeurIPS'2019 & AM3-TADAM~\cite{xing2019adaptive}$^{\dagger }$      &                          & ResNet-101         & 74.10          & 79.70          \\
      CVPR'2019    & Multi-Semantic~\cite{schwartz2019baby}$^{\dagger }$ &                          & DenseNet-121       & 76.10          & 82.90          \\
      CVPR'2020    & Hyperbolic~\cite{khrulkov2020hyperbolic}            &                          & ResNet-18          & 64.02          & 82.53          \\
      ICLR'2021    & BOIL~\cite{oh2021boil}                              &                          & 64-64-64-64        & 61.60          & 75.96          \\
      MTA'2021     & SMAVI~\cite{abdelaziz2021few}                       &                          & 64-64-64-64        & 63.54          & 76.14          \\
      TCSVT'2021   & HGNN~\cite{chen2021hierarchical}                    &                          & 64-64-64-64        & 69.43          & 87.67          \\
      PR'2021      & LMPNet~\cite{huang2021local}                        &                          & ResNet-12          & 65.59          & 68.19          \\
      WACV'2021    & Multimodal~\cite{pahde2021multimodal}$^{\dagger }$  &                          & ResNet-18          & 75.01          & 85.30          \\
      ICLRW'2021   & TFH~\cite{lazarou2021few}                           &                          & ResNet-18          & 75.83          & 88.17          \\
      \midrule
      NeurIPS'2018 & $\Delta$-encoder~\cite{schwartz2018delta}           & \multirow{6}{*}{DataAug} & ResNet-18          & 69.80          & 82.60          \\
      TIP'2019     & Dual TriNet~\cite{chen2019multi}$^{\dagger }$       &                          & ResNet-18          & 69.61          & 84.10          \\
      CVPR'2019    & f-VAEGAN-D2~\cite{xian2019f}$^{\dagger }$           &                          & ResNet-101         & 84.00          & 85.00          \\
      CVPR'2020    & AFHN~\cite{li2020adversarial}                       &                          & ResNet-18          & 70.53          & 83.95          \\
      AAAI'2020    & Deep DTN~\cite{chen2020diversity}                   &                          & ResNet-12          & 72.00          & 85.10          \\
      WACV'2021    & VI-Net~\cite{luo2021few}                            &                          & ResNet-18          & 74.76          & 86.84          \\
      \midrule
                   & \multirow{2}{*}{\textbf{DCVAE (Ours)}}              & \multirow{2}{*}{DataAug} & ResNet-12          & 77.19          & 86.97          \\
                   &                                                     &                          & ResNet-101         & \textbf{85.43} & \textbf{91.76} \\
      \bottomrule
    \end{tabular}}
  \label{cub}
  \vspace{-0.3cm}
\end{table*}

\begin{table*}[h]
  \caption{Few-shot classification accuracy (in \%) on miniImageNet dataset. ``Others'' include metric and gradient based methods, ``DataAug'' represents data augmentation based methods. $^{\dagger }$ indicates that the method is a multi-modal approach, which has leveraged the additional semantic information. The best results are in bold.}
  \vspace{-0.1cm}
  \centering
  \small
  \setlength{\tabcolsep}{3mm}{
    \begin{tabular}{lccccc}
      \toprule
      Source       & Methods                                        & Type                      & Feature Extractors & 5-way 1-shot              & 5-way 5-shot              \\

      \midrule
      ICML'2019    & TapNet~\cite{yoon2019tapnet}                   & \multirow{16}{*}{Others}  & ResNet-12          & 61.65$~\pm$ 0.15          & 76.36$~\pm$ 0.10          \\
      NeurIPS'2019 & AM3-TADAM~\cite{xing2019adaptive}$^{\dagger }$ &                           & ResNet-12          & 65.30$~\pm$ 0.49          & 78.10$~\pm$ 0.36          \\
      CVPR'2019    & MTL~\cite{sun2019meta}                         &                           & ResNet-12          & 61.20$~\pm$ 1.80          & 75.50$~\pm$ 0.80          \\
      ICLR'2019    & LEO~\cite{rusu2018meta}                        &                           & WRN-28-10          & 61.76$~\pm$ 0.08          & 77.59$~\pm$ 0.12          \\
      CVPR'2020    & Hyperbolic~\cite{khrulkov2020hyperbolic}       &                           & ResNet-18          & 59.47$~\pm$ 0.20          & 76.84$~\pm$ 0.14          \\
      CVPR'2020    & DSN~\cite{simon2020adaptive}                   &                           & ResNet-12          & 62.64$~\pm$ 0.66          & 78.83$~\pm$ 0.45          \\
      ICML'2020    & SLA-AG~\cite{lee2020self}                      &                           & ResNet-12          & 62.93$~\pm$ 0.63          & 79.63$~\pm$ 0.47          \\
      CVPR'2020    & AWGIM~\cite{guo2020attentive}                  &                           & WRN-28-10          & 63.12$~\pm$ 0.08          & 78.40$~\pm$ 0.11          \\
      IJCAI'2020   & ADM~\cite{li2020asymmetric}                    &                           & 64-64-64-64        & 54.26$~\pm$ 0.63          & 72.54$~\pm$ 0.50          \\
      ICLR'2021    & BOIL~\cite{oh2021boil}                         &                           & 64-64-64-64        & 49.61$~\pm$ 0.16          & 66.45$~\pm$ 0.37          \\
      MTA'2021     & SMAVI~\cite{abdelaziz2021few}                  &                           & 64-64-64-64        & 58.91$~\pm$ 0.85          & 74.44$~\pm$ 0.62          \\
      TCSVT'2021   & HGNN~\cite{chen2021hierarchical}               &                           & 64-64-64-64        & 60.03$~\pm$ 0.51          & 79.64$~\pm$ 0.36          \\
      arXiv'2021   & S-MoCo~\cite{majumder2021revisiting}           &                           & ResNet-18          & 59.94$~\pm$ 0.89          & 78.17$~\pm$ 0.64          \\
      ICLRW'2021   & TFH~\cite{lazarou2021few}                      &                           & ResNet-18          & 64.25$~\pm$ 0.85          & 80.10$~\pm$ 0.61          \\
      PR'2021      & LMPNet~\cite{huang2021local}                   &                           & ResNet-12          & 62.74$~\pm$ 0.11          & 80.23$~\pm$ 0.52          \\
      ICLR'2021    & ConstellationNet~\cite{xu2021attentional}      &                           & ResNet-12          & 64.89$~\pm$ 0.23          & 79.95$~\pm$ 0.17          \\

      \midrule
      NeurIPS'2018 & MetaGAN~\cite{zhang2018metagan}                & \multirow{10}{*}{DataAug} & 32-32-32-32        & 52.71$~\pm$ 0.64          & 68.63$~\pm$ 0.67          \\
      NeurIPS'2018 & $\Delta$-encoder~\cite{schwartz2018delta}      &                           & ResNet-18          & 59.90$~\pm$ n/a~          & 69.70$~\pm$ n/a~          \\
      ICCV'2019    & GCR~\cite{li2019few}                           &                           & 64-64-64-64        & 53.21$~\pm$ 0.40          & 72.34$~\pm$ 0.32          \\
      CVPR'2019    & SalNet~\cite{zhang2019few}                     &                           & ResNet-101         & 62.22$~\pm$ 0.87          & 77.95$~\pm$ 0.65          \\
      TIP'2019     & Dual TriNet~\cite{chen2019multi}$^{\dagger }$  &                           & ResNet-18          & 58.12$~\pm$ 1.37          & 76.92$~\pm$ 0.69          \\
      CVPR'2019    & IDeMe-Net~\cite{Zitian}                        &                           & ResNet-18          & 59.14$~\pm$ 0.86          & 74.63$~\pm$ 0.74          \\
      AAAI'2019    & Self-Jig~\cite{chen2019image}                  &                           & ResNet-18          & 58.80$~\pm$ 1.36          & 76.71$~\pm$ 0.72          \\
      AAAI'2020    & Deep DTN~\cite{chen2020diversity}              &                           & ResNet-12          & 63.45$~\pm$ 0.86          & 77.91$~\pm$ 0.62          \\
      CVPR'2020    & AFHN~\cite{li2020adversarial}                  &                           & ResNet-18          & 62.38$~\pm$ 0.72          & 78.16$~\pm$ 0.56          \\
      WACV'2021    & VI-Net~\cite{luo2021few}                       &                           & ResNet-18          & 61.05$~\pm$ n/a~          & 78.60$~\pm$ n/a~          \\
      \midrule
                   & \multirow{2}{*}{\textbf{DCVAE (Ours)}}         & \multirow{2}{*}{DataAug}  & ResNet-12          & 63.67$~\pm$ 0.84          & \textbf{80.59$~\pm$ 0.45} \\
                   &                                                &                           & ResNet-18          & \textbf{66.12$~\pm$ 0.25} & 78.61$~\pm$ 0.63          \\
      \bottomrule
    \end{tabular}}
  \label{mini}
  \vspace{-0.3cm}
\end{table*}

\begin{table*}[!h]
  \caption{Few-shot classification accuracy(in \%) on CIFAR-FS dataset. ``Others'' include metric and gradient based methods, ``DataAug'' represents data augmentation based methods. $^{\dagger }$ indicates that the method is a multi-modal approach, which has leveraged the additional semantic information. The best results are displayed in \textbf{boldface.}}
  \centering
  \vspace{-0.1cm}
  \small
  \setlength{\tabcolsep}{3mm}{
    \begin{tabular}{lccccc}
      \toprule
      Source       & Methods                                       & Type                     & Feature Extractors & 5-way 1-shot            & 5-way 5-shot            \\

      \midrule
      NeurIPS'2017 & Prototypical Net~\cite{snell2017prototypical} & \multirow{13}{*}{Others} & ResNet-12          & 72.2$~\pm$ 0.7          & 83.5$~\pm$ 0.5          \\
      TIP'2019     & Two Stage~\cite{das2019two}                   &                          & 64-64-64-64        & 67.1$~\pm$ 0.3          & 81.6$~\pm$ 0.3          \\
      ICCV'2019    & Shot-Free~\cite{ravichandran2019few}          &                          & ResNet-12          & 69.2$~\pm$ n/a          & 84.7$~\pm$ n/a          \\
      CVPR'2019    & MetaOptNet-RR~\cite{lee2019meta}              &                          & ResNet-12          & 72.6$~\pm$ 0.7          & 84.3$~\pm$ 0.5          \\
      CVPR'2019    & MetaOptNet-SVM~\cite{lee2019meta}             &                          & ResNet-12          & 72.0$~\pm$ 0.7          & 84.2$~\pm$ 0.5          \\
      ECCV'2020    & BAS~\cite{kim2020model}                       &                          & ResNet-12          & 73.5$~\pm$ 0.9          & 85.5$~\pm$ 0.6          \\
      CVPR'2020    & LR+ICI~\cite{wang2020instance}                &                          & ResNet-12          & 73.9$~\pm$ n/a          & 84.1$~\pm$ n/a          \\
      CVPR'2020    & DSN-MR~\cite{simon2020adaptive}               &                          & ResNet-12          & 75.6$~\pm$ 0.9          & 86.2$~\pm$ 0.6          \\
      ICML'2020    & SLA-AG~\cite{lee2020self}                     &                          & ResNet-12          & 74.6$~\pm$ 0.7          & 86.8$~\pm$ 0.5          \\
      ECCV'2020    & RFS-distill~\cite{tian2020rethinking}         &                          & ResNet-12          & 73.9$~\pm$ 0.8          & 86.9$~\pm$ 0.5          \\
      ICLR'2021    & ConstellationNet~\cite{xu2021attentional}     &                          & ResNet-12          & 75.4$~\pm$ 0.2          & 86.8$~\pm$ 0.2          \\
      arXiv'2021   & S-MoCo~\cite{majumder2021revisiting}          &                          & ResNet-18          & 69.2$~\pm$ 0.9          & 85.3$~\pm$ 0.7          \\
      ICLRW'2021   & TFH~\cite{lazarou2021few}                     &                          & ResNet-18          & 73.8$~\pm$ 0.8          & 85.9$~\pm$ 0.6          \\

      \midrule
      NeurIPS'2018 & $\Delta$-encoder~\cite{schwartz2018delta}     & \multirow{4}{*}{DataAug} & ResNet-18          & 66.7$~\pm$ n/a          & 79.8$~\pm$ n/a          \\
      TIP'2019     & Dual TriNet~\cite{chen2019multi}$^{\dagger }$ &                          & ResNet-18          & 63.4$~\pm$ 0.6          & 78.4$~\pm$ 0.6          \\
      CVPR'2020    & AFHN~\cite{li2020adversarial}                 &                          & ResNet-18          & 68.3$~\pm$ 0.9          & 81.5$~\pm$ 0.9          \\
      AAAI'2020    & Deep DTN~\cite{chen2020diversity}             &                          & ResNet-12          & 71.5$~\pm$ n/a          & 82.8$~\pm$ n/a          \\

      \midrule
                   & \multirow{2}{*}{\textbf{DCVAE(Ours)}}         & \multirow{2}{*}{DataAug} & ResNet-12          & \textbf{76.1$~\pm$ 0.6} & \textbf{87.0$~\pm$ 0.3} \\
                   &                                               &                          & ResNet-18          & 74.9$~\pm$ 0.4          & 86.5$~\pm$ 0.5          \\
      \bottomrule
    \end{tabular}}
  \label{cifar}
  \vspace{-0.3cm}
\end{table*}

\vspace{-0.1cm}
\section{Experiments}
In this section, we conduct the experiments on three popular benchmarks, namely CIFAR-FS, CUB, miniImageNet, for evaluating our work in comparison with the recent state-of-the-art methods.

\vspace{-0.3cm}
\subsection{Datasets and Settings}
The miniImageNet~\cite{vinyals2016matching} consists of 100 classes randomly sampled from the ILSVRC2012 dataset~\cite{russakovsky2015imagenet}, each class has 600 images with size 84$\times$84 pixels. Following the same split proposed by \cite{snell2017prototypical}, we take 64 categories for training, 16 for validation and 20 for test, respectively. The CUB-200-2011(CUB) dataset is proposed by \cite{wah2011caltech}, which is a popular benchmark dataset originally for fine-grained classification. It contains 11,788 images from 200 categories of birds annotated with 312 attributes. We follow the training/validation/testing split proposed by \cite{xian2017zero} for evaluation. CIFAR-FS is derived from the standard CIFAR-100 dataset~\cite{krizhevsky2009learning}, which randomly divides 100 classes into 64 training classes, 16 validation classes and 20 testing classes.

We use the 300-dimensional word embedding obtained by GloVe\cite{pennington2014glove} via feeding the class name of each category as the semantic representation in the miniImageNet. With regard to the CUB dataset, we take advantage of the publicly available attributes provided by \cite{xian2017zero} as the semantic representation of class.

Following the previous approaches~\cite{li2020adversarial,chen2020diversity}, we evaluate 5-way 1-shot and 5-way 5-shot classification tasks. During the test stage, for each task $\mathcal{T}$, we use the support set $\mathcal{S_{\mathcal{T}}}$ for fine-tuning the trained model. We respectively conduct 50 and 100 times of fine-tuning in 5-way 1-shot and 5-way 5-shot tasks. Finally, we synthesize 100 fake features for each class and adopt the Euclidean distance based $k$-NN classifier for classification. Here, we set $k=5$. Following the usual settings, our model is evaluated over 600 episodes with 15 test samples from each category. Our project is implemented in PyTorch and will be released in the future.

\vspace{-0.3cm}
\subsection{Few-Shot Learning}
\vspace{-0.1cm}
Table~\ref{cub} tabulates the few-shot classification accuracy of different compared methods on CUB dataset. From the results, it is clear that DCVAE achieves the best performances with the significant advantages in all experiments. For example, the performance gains of our method over the second best one are 1.43\% and 6.76\% in 5-way 1-shot and 5-way 5-shot tasks respectively in the case of ResNet-101. $\Delta$-encoder, AFHN, VI-Net and Deep DTN are the single-modal data augmentation-based approaches. DCVAE gets 15.63\%, 14.90\%, 10.67\% and 13.43\% more performance gains over these four methods respectively in 5-way 1-shot task. We attribute these significant performance gains to the fact that these data augmentation approaches only consider the single-modal condition during sample generation while our work considers the semantic and visual conditions both. Multimodal~\cite{pahde2021multimodal}, AM3-TADAM~\cite{xing2019adaptive}, Multi-Semantic~\cite{schwartz2019baby} are the multi-modal approaches which leverage the extra semantic information like ours. However, DCVAE also gets 10.42\%, 11.33\% and 9.33\% accuracy improvements over these methods respectively in 5-way 1-shot task. Dual TriNet~\cite{chen2019multi} and f-VAEGAN-D2~\cite{xian2019f} are the multi-modal data augmentation-based methods as same as DCVAE. DCVAE still shows its prominent advantages over these methods in both 5-way 1-shot and 5-way 5-shot task. Moreover, the ResNet-12 Feature Extractor-based DCVAE outperforms all the compared methods with ResNet-12 and four-layer CNN backbones, and also performs much better than most of the compared approaches with a heavier backbone network, e.g., Multi-Semantic (DenseNet-121)~\cite{schwartz2019baby} ,AM3-TADAM (ResNet-101)~\cite{xing2019adaptive}, Dual TriNet (ResNet-18)~\cite{chen2019multi} and so on.

Table~\ref{mini} records the few-shot classification accuracies of different methods on miniImageNet. DCVAE achieves very promising performances in comparison with the state-of-the-art approaches.
It outperforms all compared methods with ResNet-18 backbone in 5-way 1-shot task and with ResNet-12 backbone in 5-way 5-shot task respectively. Deep DTN~\cite{chen2020diversity} is the best performed data augmentation-based compared method on miniImageNet. DCVAE outperforms it with 2.67\% and 2.68\% accuracy improvements in 5-way 1-shot and 5-way 5-shot tasks respectively. AM3-TADAM~\cite{xing2019adaptive} is the best performed multi-modal few-shot learning approach on miniImageNet. DCVAE still gains more 0.82\% and 2.49\% accuracies in 5-way 1-shot and 5-way 5-shot tasks respectively.

Table~\ref{cifar} reports the few-shot learning results on CIFAR-FS. Similar to the observations on CUB and miniImageNet dataset, DCVAE is still the best performed approach on CIFAR-FS dataset, and the performance advantages of DCVAE over other compared approaches on CIFAR-FS dataset are even more significant than the ones on miniImageNet dataset. Dual TriNet~\cite{chen2019multi} is the only multi-modal data augmentation-based methods among all compared approaches. DCVAE gains more 12.7\% and 8.6\% accuracies over it in 5-way 1-shot and 5-way 5-shot tasks respectively. Deep DTN~\cite{chen2020diversity} is the best performed data augmentation-based compared method. The accuracy improvements of our method over Deep DTN in 5-way 1-shot and 5-way 5-shot tasks are 4.6\% and 4.2\% respectively.

In summary, all these experimental results demonstrate that DCVAE is a promising data augmentation approach for few-shot learning via properly exploiting the complementary of different modality information.

\begin{table*}[htb]
  \caption{The 5-way few-shot classification accuracy(in \%) of DCVAE under different absence ratios ($\eta_{m}$) of visual and semantic modalities on miniImageNet, CUB and CIFAR-FS datasets. $\eta_{s} + \eta_{v} \leq 100\%$ for making sure that each sample should be represented by at least one modality.}
  \centering
  \subtable[1-shot on miniImageNet]{
    \setlength{\tabcolsep}{0.6mm}{
      \begin{tabular}{lcccccc}
        \toprule
        \diagbox{$\eta_{v}$}{$\eta_{s}$} & 0\%   & 20\%  & 40\%  & 60\%  & 80\%  & 100\% \\

        \midrule
        \multicolumn{1}{c}{0\%}          & 66.12 & 64.41 & 63.25 & 60.79 & 59.53 & 59.42 \\
        \midrule
        \multicolumn{1}{c}{20\%}         & 65.73 & 63.56 & 62.16 & 59.56 & 58.33 & -     \\
        \midrule
        \multicolumn{1}{c}{40\%}         & 63.68 & 61.98 & 59.78 & 57.51 & -     & -     \\
        \midrule
        \multicolumn{1}{c}{60\%}         & 61.21 & 59.67 & 57.54 & -     & -     & -     \\
        \midrule
        \multicolumn{1}{c}{80\%}         & 60.29 & 58.34 & -     & -     & -     & -     \\
        \midrule
        \multicolumn{1}{c}{100\%}        & 59.75 & -     & -     & -     & -     & -     \\
        \bottomrule
      \end{tabular}
    }
  }
  \subtable[1-shot on CIFAR-FS]{
    \setlength{\tabcolsep}{0.6mm}{
      \begin{tabular}{lcccccc}
        \toprule
        \diagbox{$\eta_{v}$}{$\eta_{s}$} & 0\%   & 20\%  & 40\%  & 60\%  & 80\%  & 100\% \\

        \midrule
        \multicolumn{1}{c}{0\%}          & 76.10 & 74.37 & 73.65 & 73.37 & 73.08 & 73.15 \\
        \midrule
        \multicolumn{1}{c}{20\%}         & 75.46 & 73.81 & 73.09 & 72.46 & 71.4  & -     \\
        \midrule
        \multicolumn{1}{c}{40\%}         & 73.72 & 71.93 & 70.99 & 70.15 & -     & -     \\
        \midrule
        \multicolumn{1}{c}{60\%}         & 73.48 & 70.19 & 69.47 & -     & -     & -     \\
        \midrule
        \multicolumn{1}{c}{80\%}         & 71.65 & 69.97 & -     & -     & -     & -     \\
        \midrule
        \multicolumn{1}{c}{100\%}        & 70.18 & -     & -     & -     & -     & -     \\
        \bottomrule
      \end{tabular}
    }
  }
  \subtable[1-shot on CUB]{
    \setlength{\tabcolsep}{0.6mm}{
      \begin{tabular}{lcccccc}
        \toprule
        \diagbox{$\eta_{v}$}{$\eta_{s}$} & 0\%   & 20\%  & 40\%  & 60\%  & 80\%  & 100\% \\

        \midrule
        \multicolumn{1}{c}{0\%}          & 85.43 & 82.72 & 81.57 & 78.38 & 77.53 & 80.65 \\
        \midrule
        \multicolumn{1}{c}{20\%}         & 83.75 & 81.97 & 79.61 & 77.86 & 76.46 & -     \\
        \midrule
        \multicolumn{1}{c}{40\%}         & 81.19 & 79.04 & 76.94 & 75.59 & -     & -     \\
        \midrule
        \multicolumn{1}{c}{60\%}         & 78.72 & 74.69 & 74.58 & -     & -     & -     \\
        \midrule
        \multicolumn{1}{c}{80\%}         & 78.04 & 74.47 & -     & -     & -     & -     \\
        \midrule
        \multicolumn{1}{c}{100\%}        & 77.93 & -     & -     & -     & -     & -     \\
        \bottomrule
      \end{tabular}
    }
  }
  \subtable[5-shot on miniImageNet]{
    \setlength{\tabcolsep}{0.6mm}{
      \begin{tabular}{lcccccc}
        \toprule
        \diagbox{$\eta_{v}$}{$\eta_{s}$} & 0\%   & 20\%  & 40\%  & 60\%  & 80\%  & 100\% \\

        \midrule
        \multicolumn{1}{c}{0\%}          & 78.61 & 76.15 & 75.86 & 75.63 & 75.38 & 75.58 \\
        \midrule
        \multicolumn{1}{c}{20\%}         & 72.24 & 70.81 & 69.39 & 71.13 & 71.79 & -     \\
        \midrule
        \multicolumn{1}{c}{40\%}         & 66.73 & 64.47 & 65.96 & 67.32 & -     & -     \\
        \midrule
        \multicolumn{1}{c}{60\%}         & 63.15 & 62.85 & 64.89 & -     & -     & -     \\
        \midrule
        \multicolumn{1}{c}{80\%}         & 60.26 & 62.30 & -     & -     & -     & -     \\
        \midrule
        \multicolumn{1}{c}{100\%}        & 59.83 & -     & -     & -     & -     & -     \\
        \bottomrule
      \end{tabular}
    }
  }
  \subtable[5-shot on CIFAR-FS]{
    \setlength{\tabcolsep}{0.6mm}{
      \begin{tabular}{lcccccc}
        \toprule
        \diagbox{$\eta_{v}$}{$\eta_{s}$} & 0\%   & 20\%  & 40\%  & 60\%  & 80\%  & 100\% \\

        \midrule
        \multicolumn{1}{c}{0\%}          & 87.00 & 86.03 & 85.72 & 84.98 & 84.06 & 84.51 \\
        \midrule
        \multicolumn{1}{c}{20\%}         & 85.43 & 83.87 & 82.39 & 81.07 & 82.75 & -     \\
        \midrule
        \multicolumn{1}{c}{40\%}         & 82.25 & 80.63 & 80.16 & 79.29 & -     & -     \\
        \midrule
        \multicolumn{1}{c}{60\%}         & 80.95 & 78.91 & 77.27 & -     & -     & -     \\
        \midrule
        \multicolumn{1}{c}{80\%}         & 76.92 & 73.94 & -     & -     & -     & -     \\
        \midrule
        \multicolumn{1}{c}{100\%}        & 70.64 & -     & -     & -     & -     & -     \\
        \bottomrule
      \end{tabular}
    }
  }
  \subtable[5-shot on CUB]{
    \setlength{\tabcolsep}{0.6mm}{
      \begin{tabular}{lcccccc}
        \toprule
        \diagbox{$\eta_{v}$}{$\eta_{s}$} & 0\%   & 20\%  & 40\%  & 60\%  & 80\%  & 100\% \\

        \midrule
        \multicolumn{1}{c}{0\%}          & 91.76 & 90.36 & 89.97 & 89.41 & 89.08 & 89.22 \\
        \midrule
        \multicolumn{1}{c}{20\%}         & 90.82 & 89.18 & 88.95 & 87.55 & 88.47 & -     \\
        \midrule
        \multicolumn{1}{c}{40\%}         & 89.38 & 87.83 & 86.36 & 87.19 & -     & -     \\
        \midrule
        \multicolumn{1}{c}{60\%}         & 86.71 & 86.07 & 85.95 & -     & -     & -     \\
        \midrule
        \multicolumn{1}{c}{80\%}         & 82.14 & 84.78 & -     & -     & -     & -     \\
        \midrule
        \multicolumn{1}{c}{100\%}        & 78.16 & -     & -     & -     & -     & -     \\
        \bottomrule
      \end{tabular}
    }
  }
  \vspace{-0.5cm}
  \label{mm}
\end{table*}

\vspace{-0.3cm}
\subsection{Partial Modality Absent Few-Shot Learning}
In this section, we conduct several experiments on all three benchmarks to validate the partial modality absent few-shot learning abilities of DCVAE. The partial modality absent few-shot learning means that some examples lack the completed modality information as the cases shown in Figure~\ref{modal}(d)-(f).

In these experiments, we randomly remove some modalities of samples to simulate different modality absence scenarios, and define the modality absence ratio to measure the severity of modality absence with respect to each type of modality. The absence ratio of modality $m$ is denoted as $\eta_{m} = \frac{N_{m}}{N} $ where $N_{m}$ is the number of samples that the modality $m$ is absence, and $N$ is the number of samples in a few-shot task. According to the absence situations of semantic and visual modality, we can compute $\eta_{s}$ and $\eta_{v}$ respectively.

Table~\ref{mm} records the performances of DCVAE under different combinations of $\eta_{s}$ and $\eta_{v}$. Note, each sample needs to be represented by at least one modality, so $\eta_{s} + \eta_{v} \leq 100\%$. Generally speaking, the common few-shot learning approaches are theoretically able to handle some modality absence situations that all samples share at least one same type of modality information, e.g., $\eta_{s}= 0\%$ or $\eta_{v}= 0\% $, via considering these cases as a single modality few-shot learning problem (only visual information is available) or a zero-shot learning problem (only semantic information is available) for solution. However, it is clear that such a way directly ignores the other existing modality information, and is unable to handle the more general modality absence situation that ($\eta_{s}> 0\%~\&~\eta_{v}>0\%$). The experimental results on all benchmarks show that the accuracies are decreased along with $\eta$ increasing as shown in each column and each row, and the full modality ($\eta_{s}=0\%~\&~\eta_{v}=0\%$) case consistently performs better than the single modality cases ($\eta_{s}=100\%~\&~\eta_{v}=0\%$ or $\eta_{s}=0\%~\& ~\eta_{v}=100\%$). These observations all reveal two facts. One is that DCVAE is able to work in all scenarios shown in Figure~\ref{modal}, including single-modal, multi-modal, modality absence and cross-modal cases. The other is that DCVAE is able to sufficiently exploit information of all existing modality among samples for improving few-shot classification performance in the general modality absence scenario. In short, DCVAE provides a more general, elegant, and proper way to address the few-shot learning task under different modality absence scenarios.

\begin{figure*}[h]
  \centering
  \subfigure[CUB]{
    \includegraphics[scale=0.235]{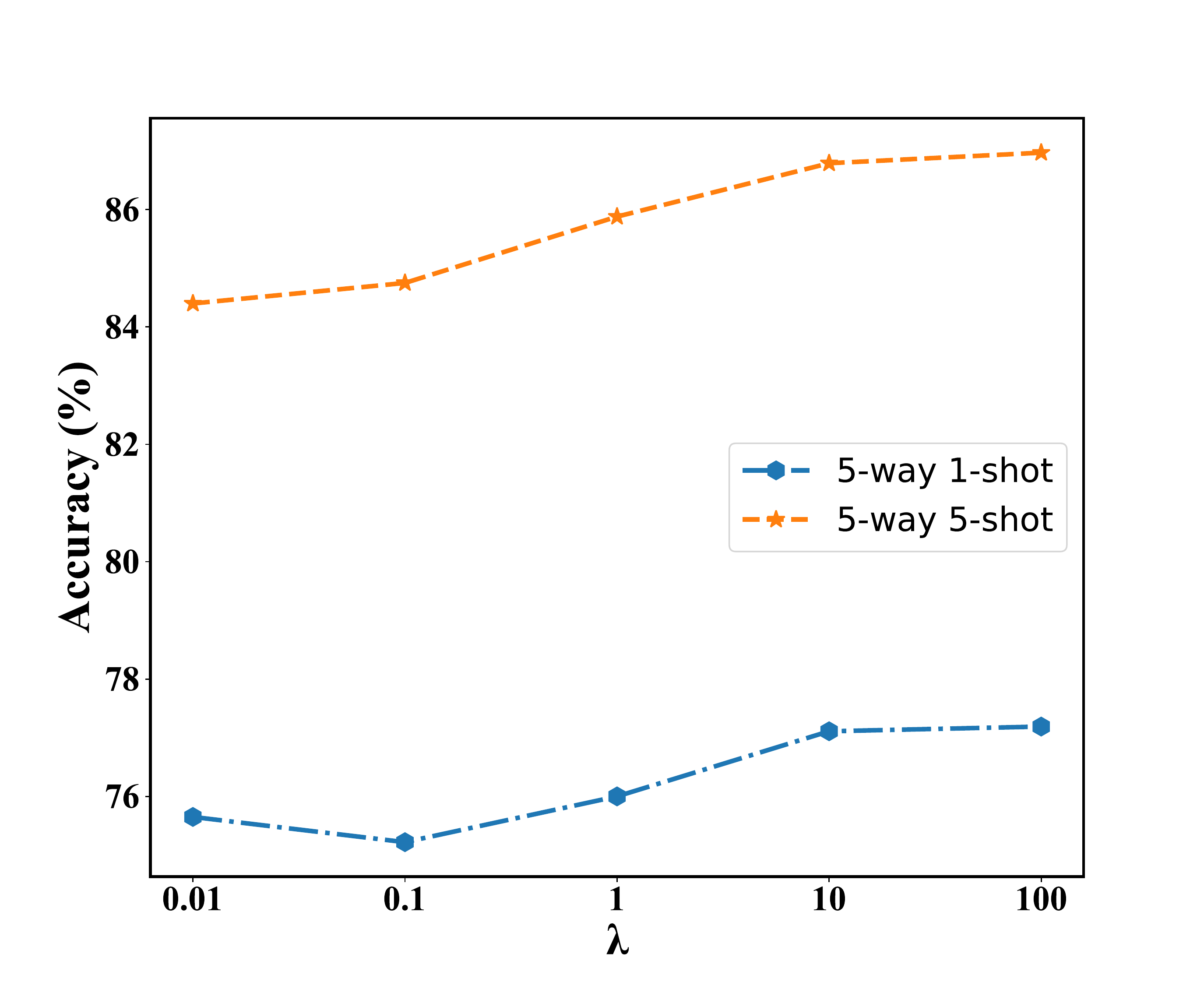}
  }
  \subfigure[CIFAR-FS]{
    \includegraphics[scale=0.235]{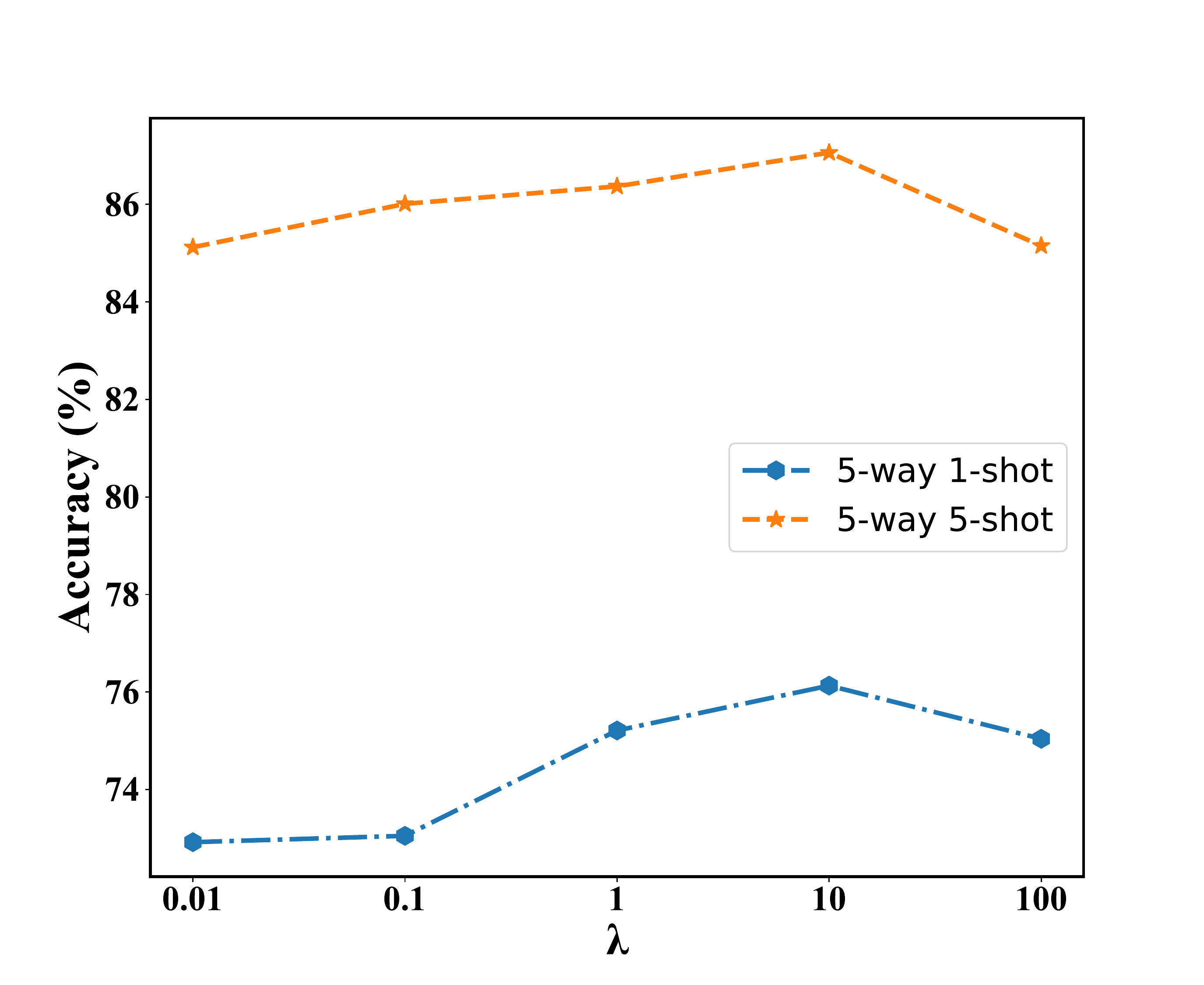}
  }
  \subfigure[miniImageNet]{
    \includegraphics[scale=0.235]{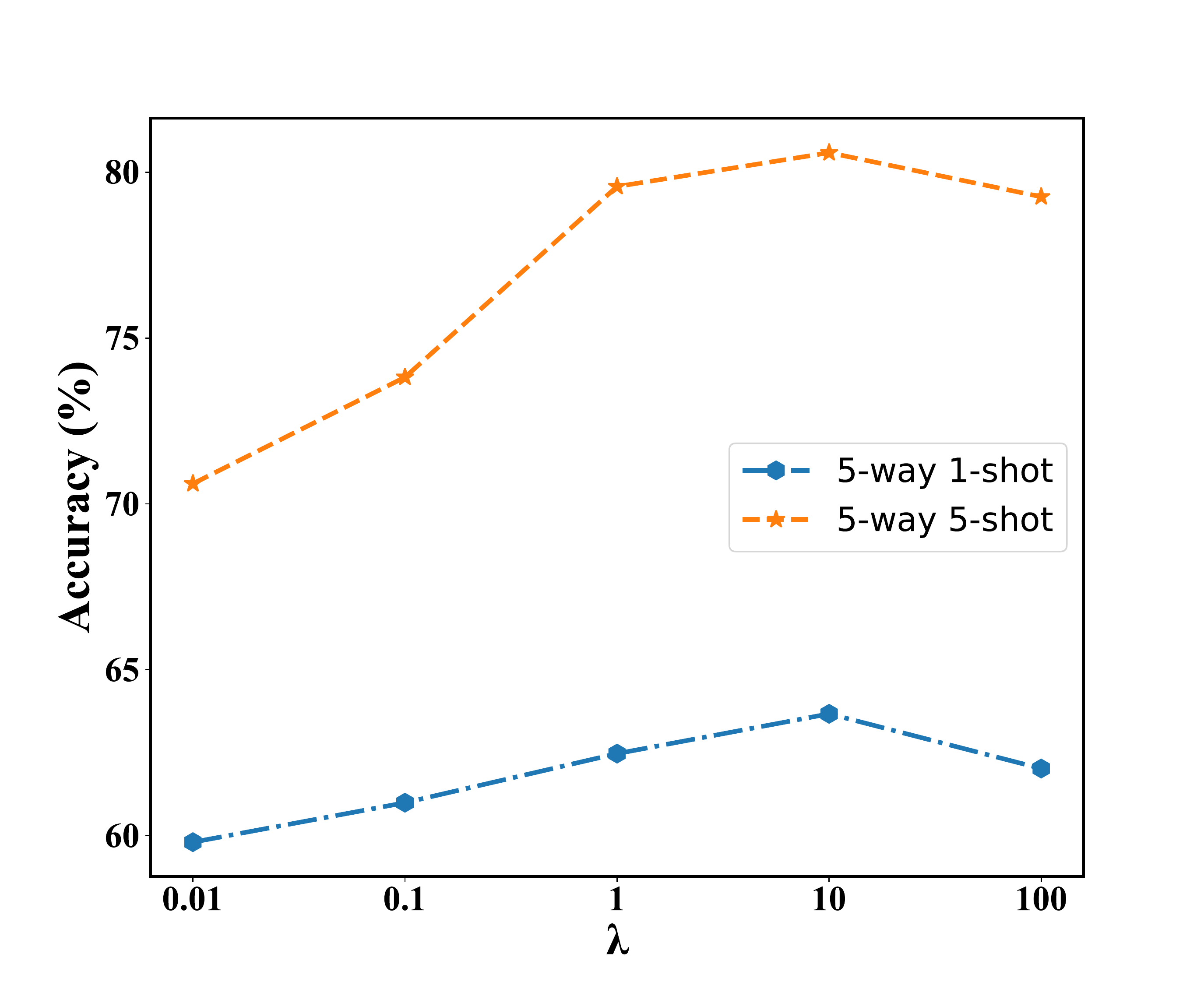}
  }
  \vspace{-0.2cm}
  \caption{Effect of Hyper-parameters $\lambda$ on three benchmarks. By varying the value of $\lambda$ from \{0.01, 0.1, 1, 10, 100\}, we can explore the effect of hyper-parameter on the performance of the model.}
  \label{hp}
  \vspace{-0.5cm}
\end{figure*}

\vspace{-0.2cm}
\subsection{Parameter Sensitivity Studies}
\subsubsection{Effect of Hyper-parameters $\lambda$}
Hyper-parameter $\lambda$ is used for reconciling the different losses in our model. By varying the value of $\lambda$ from \{0.01, 0.1, 1, 10, 100\}, we can find the influence of $\lambda$ on the performance of the proposed model. To this end, we evaluate our method on all three datasets for 5-way 1-shot and 5-way 5-shot tasks.

As shown in Figure~\ref{hp}, we observe that the best performance is achieved at $\lambda=100$ in the 1-shot task of the CUB. In the 5-shot task, the model performance decreases slightly at the beginning and then increases along with the increase of $\lambda$, again achieving the best performance at $\lambda=100$. Furthermore, the growth trends of model performance on CIFAR-FS and miniImageNet are very similar. Their best $\lambda$ are 10.

\subsubsection{Effect of the Number of $k$ in $k$-NN} We conduct several experiments to study the impacts of different $k$ in the $k$-NN classifier to the performance of DCVAE. Table~\ref{knn} tabulates the 5-way classification performances of our model under different $k$ on miniImageNet. From the results, it is not hard to find that our method is quite insensitive to the setting of $k$, and we suggest $k=5$.

\begin{table}[htbp]
  \caption{The 5-way classification accuracy of DCVAE under different $k \in \{1,3,5,7,9\}$ on miniImageNet. The best result is in bold.}
  \centering
  \begin{tabular}{lccccc}
    \toprule
    \multirow{2.5}{*}{Model} & \multicolumn{5}{c}{5-way Accuracy(\%)}                                          \\
    \cmidrule{2-6}
                             & k = 1                                  & k = 3 & k = 5          & k = 7 & k = 9 \\
    \midrule
    1-shot                   & 65.70                                  & 66.11 & \textbf{66.12} & 66.07 & 66.10 \\
    5-shot                   & 76.62                                  & 78.53 & \textbf{78.61} & 78.60 & 78.58 \\
    \bottomrule
  \end{tabular}
  \label{knn}
  \vspace{-0.2cm}
\end{table}

\begin{table}[htp]
  \caption{Effect of the number of the final synthetic features on the miniImageNet dataset. The best results are in \textbf{bold.}}
  \centering
  \setlength{\tabcolsep}{0.9mm}{
    \begin{tabular}{lccccccc}
      \toprule
      Setting & n=0   & n=50  & n=100          & n=200 & n=300 & n=400 & n=500          \\
      \midrule
      1-shot  & 56.13 & 65.31 & \textbf{66.12} & 66.04 & 66.02 & 65.92 & 66.10          \\
      5-shot  & 68.45 & 78.51 & 78.61          & 78.53 & 78.50 & 78.49 & \textbf{78.65} \\
      \bottomrule
    \end{tabular}}
  \label{ns}
  \vspace{-0.3cm}
\end{table}

\subsubsection{Effect of the Number of Synthetic Features}
In this subsection, we analyze how the final synthetic feature amount complemented in support set affects test accuracy on miniImageNet, as shown in Table \ref{ns}. We achieve at least 10\% accuracy improvement via complementing the features generated by our DCVAE to the original support set in both 5-way 1-shot and 5-way 5-shot tasks respectively. From the experimental results, it is not hard to find that the performance is quite insensitive to the setting of the complemented synthesized sample amount $n$ when $n \geq 0$. These observations reveal that the augmented sample prominently alleviate the data scarcity, and 50 synthesized samples are enough for complementing the support set to train a reliable classification model. Here, we set $n=100$.


\begin{table}[htbp]
  \caption{FSL results (in \%) on miniImageNet with combinations of loss functions and different modules. ``w/o'' and ``w/'' indicate without and with. ``$x_{s}$'', ``$x_{v}$'' and ``$\hat{x}$'' respectively denotes the semantic-based synthetic feature, visual-based synthetic feature and the final synthetic feature.}
  \centering
  \begin{tabular}{lccc}
    \toprule
    Loss Function                                                                                                 & Type      & 1-shot & 5-shot \\
    \midrule
    $\mathcal{L_{\mathsf{b} \mathsf{c} \mathsf{v} \mathsf{a} \mathsf{e} }}$ w/o DFAM                              & $x_{s}$   & 63.28  & 76.48  \\
    $\mathcal{L_{\mathsf{b} \mathsf{c} \mathsf{v} \mathsf{a} \mathsf{e} }}$ w/o DFAM                              & $x_{v}$   & 59.04  & 74.92  \\
    $\mathcal{L_{\mathsf{b} \mathsf{c} \mathsf{v} \mathsf{a} \mathsf{e} }}$ w/ DFAM                               & $\hat{x}$ & 65.30  & 77.79  \\
    $\mathcal{L_{\mathsf{b} \mathsf{c} \mathsf{v} \mathsf{a} \mathsf{e} }} + \mathcal{L}_{ts}$                    & $\hat{x}$ & 65.38  & 77.85  \\
    $\mathcal{L_{\mathsf{b} \mathsf{c} \mathsf{v} \mathsf{a} \mathsf{e} }} + \mathcal{L}_{ts} + \mathcal{L}_{rc}$ & $\hat{x}$ & 65.67  & 78.03  \\
    DCVAE w/ CCC                                                                                                  & $\hat{x}$ & 66.01  & 78.35  \\
    \bottomrule
  \end{tabular}
  \label{as}
  \vspace{-0.3cm}
\end{table}

\subsection{Ablation Study}
\subsubsection{Influences of Different Losses}
We study the influences of different loss terms in Eq.~\ref{final} by progressively integrating them into our proposed model. To this end, we evaluate our method on the miniImageNet for 5-way setting with the ResNet18 backbone and the results are reported in Table~\ref{as}. We observe that the integration of all three terms in our loss consistently outperforms any other configuration, it boosts the performance from 65.30\% to 66.01\% in 1-shot task. We also find that twin similarity loss $\mathcal{L}_{ts}$, only marginally improved the performance of our model. We attribute this to the fact that the twin similarity loss and the reconstruction loss share some overlapping functions on the constraints of generated features, which all tend to make the generated features as similar as possible to the original features and their twins.

\begin{figure*}[htp]
  \centering
  \subfigure[CVAE(v)(dis: 0.064)]{\label{aa}
    \includegraphics[scale=0.37]{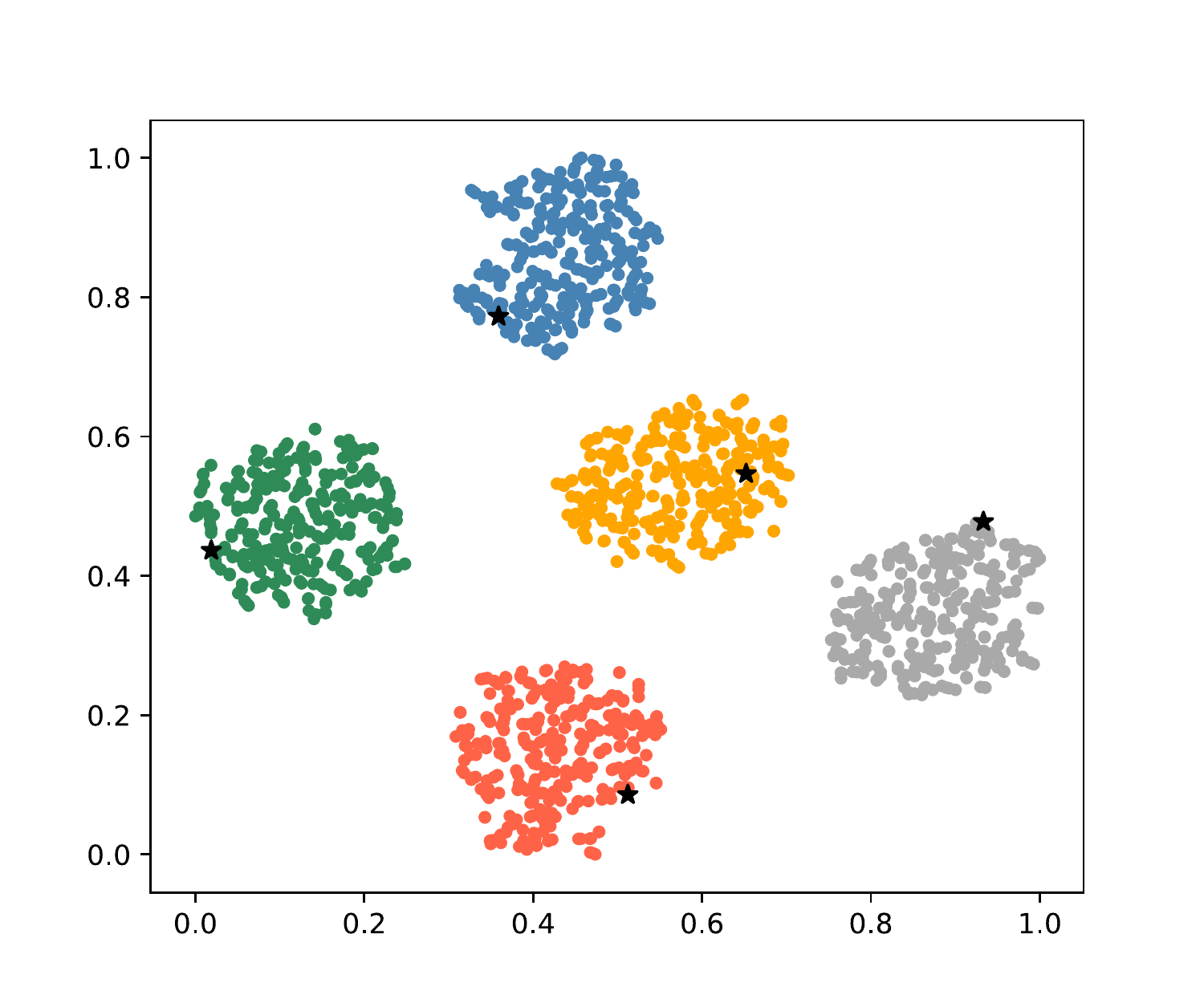}
  }
  \subfigure[CVAE(s)(dis: 0.027)]{\label{bb}
    \includegraphics[scale=0.37]{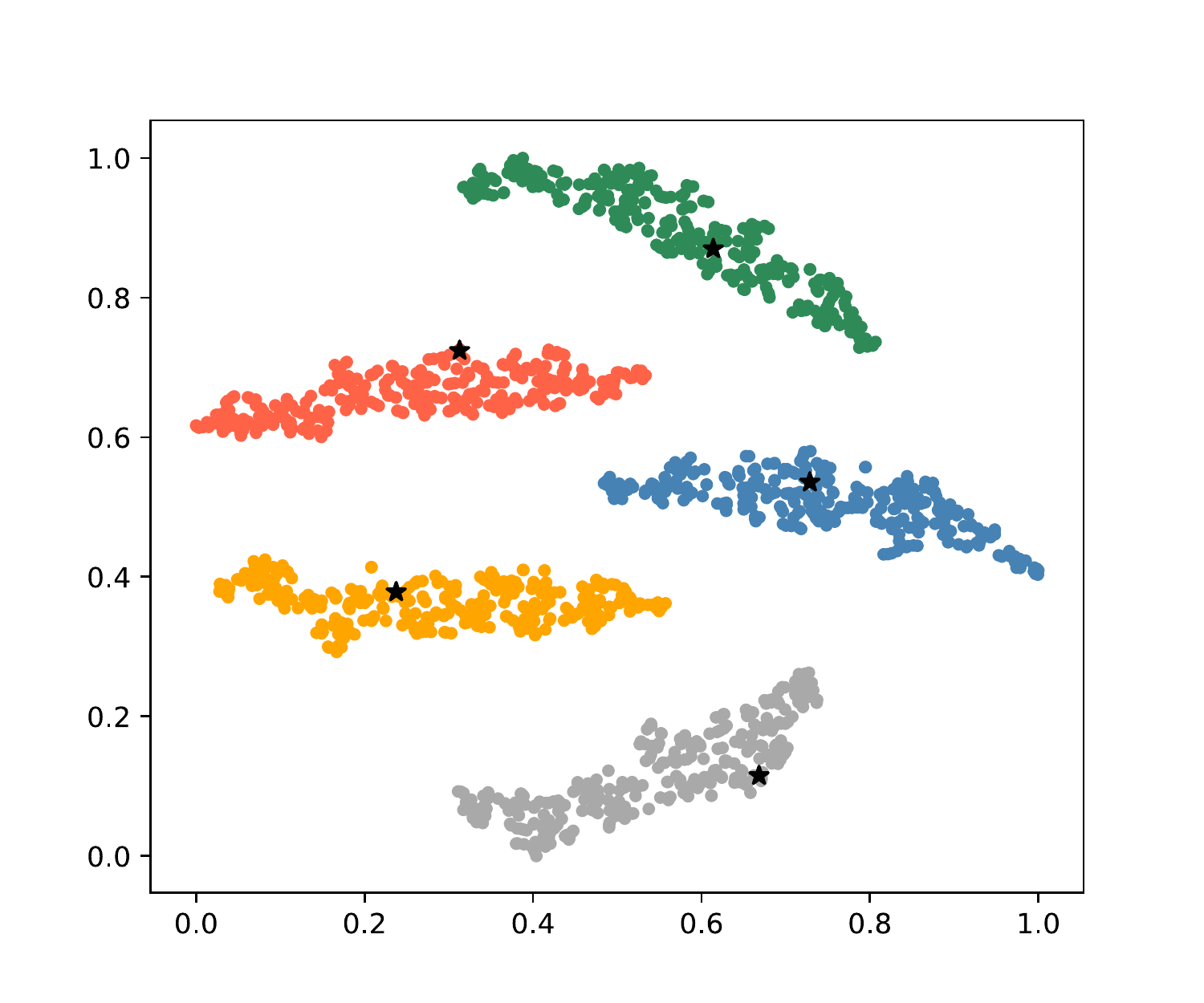}
  }
  \subfigure[CVAE(s;v)(dis: 0.039)]{\label{cc}
    \includegraphics[scale=0.37]{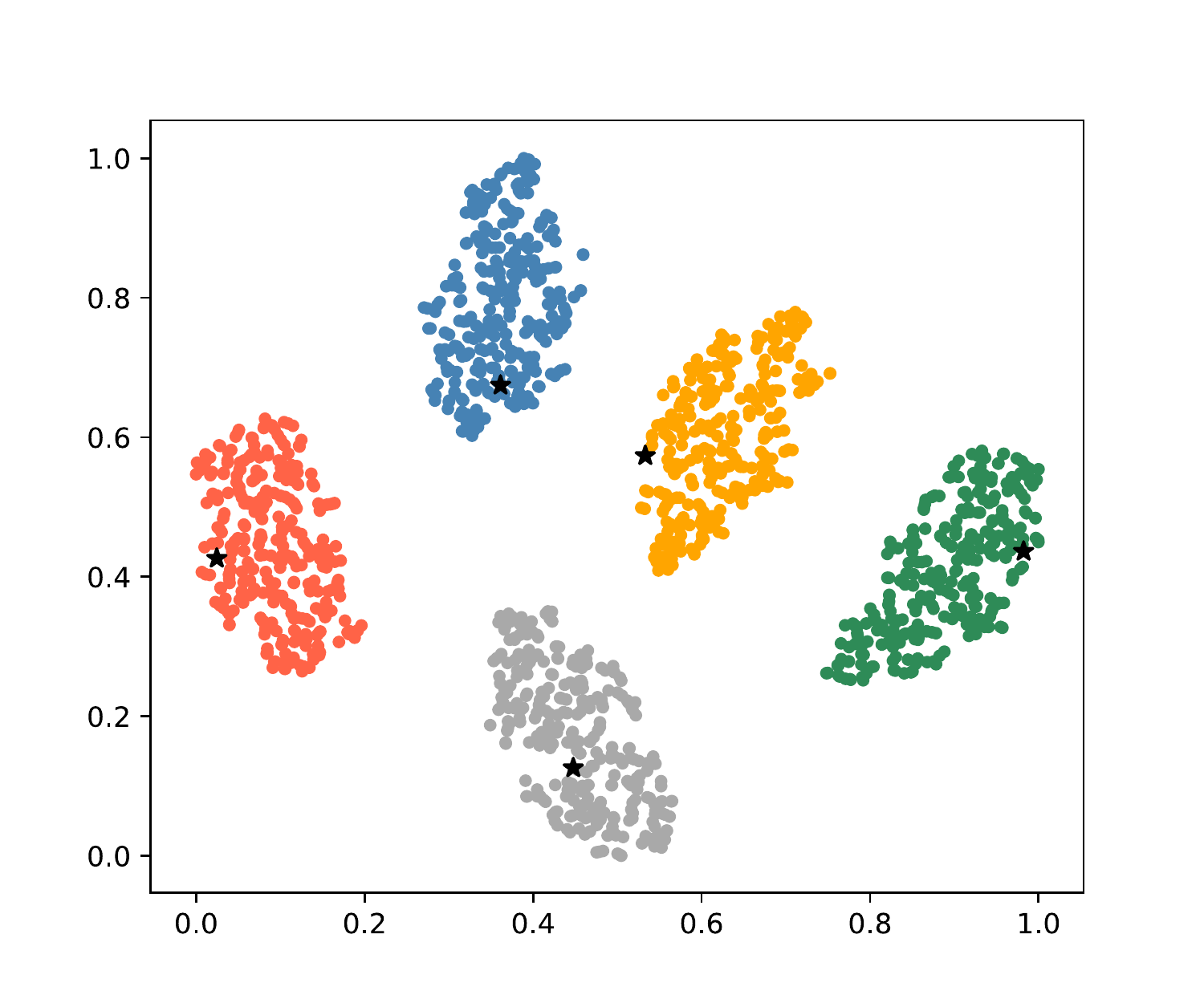}
  }
  \subfigure[DCVAE($x_{v}$)(dis: 0.036)]{\label{dd}
    \includegraphics[scale=0.37]{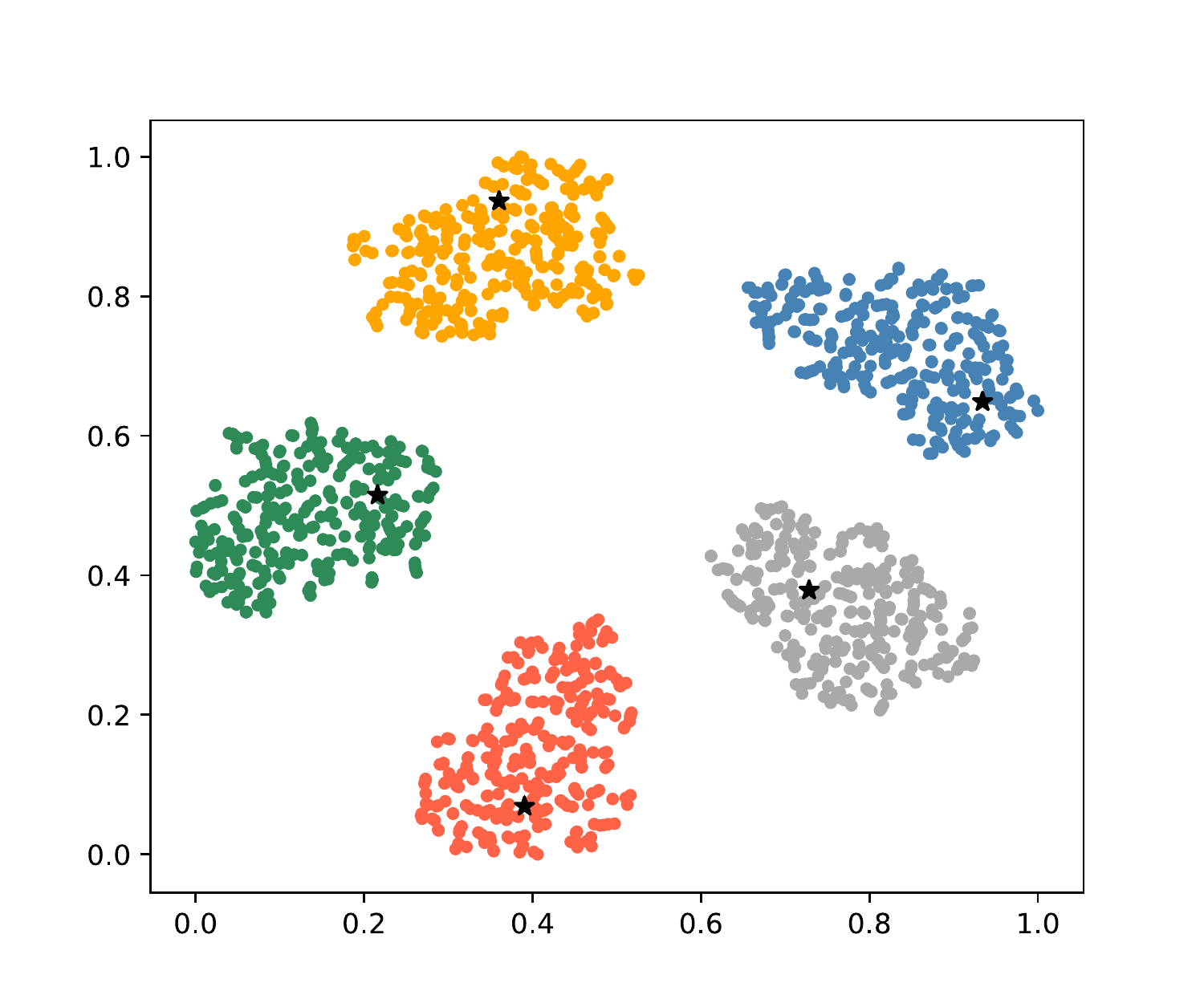}
  }
  \subfigure[DCVAE($x_{s}$)(dis: 0.008)]{\label{ee}
    \includegraphics[scale=0.37]{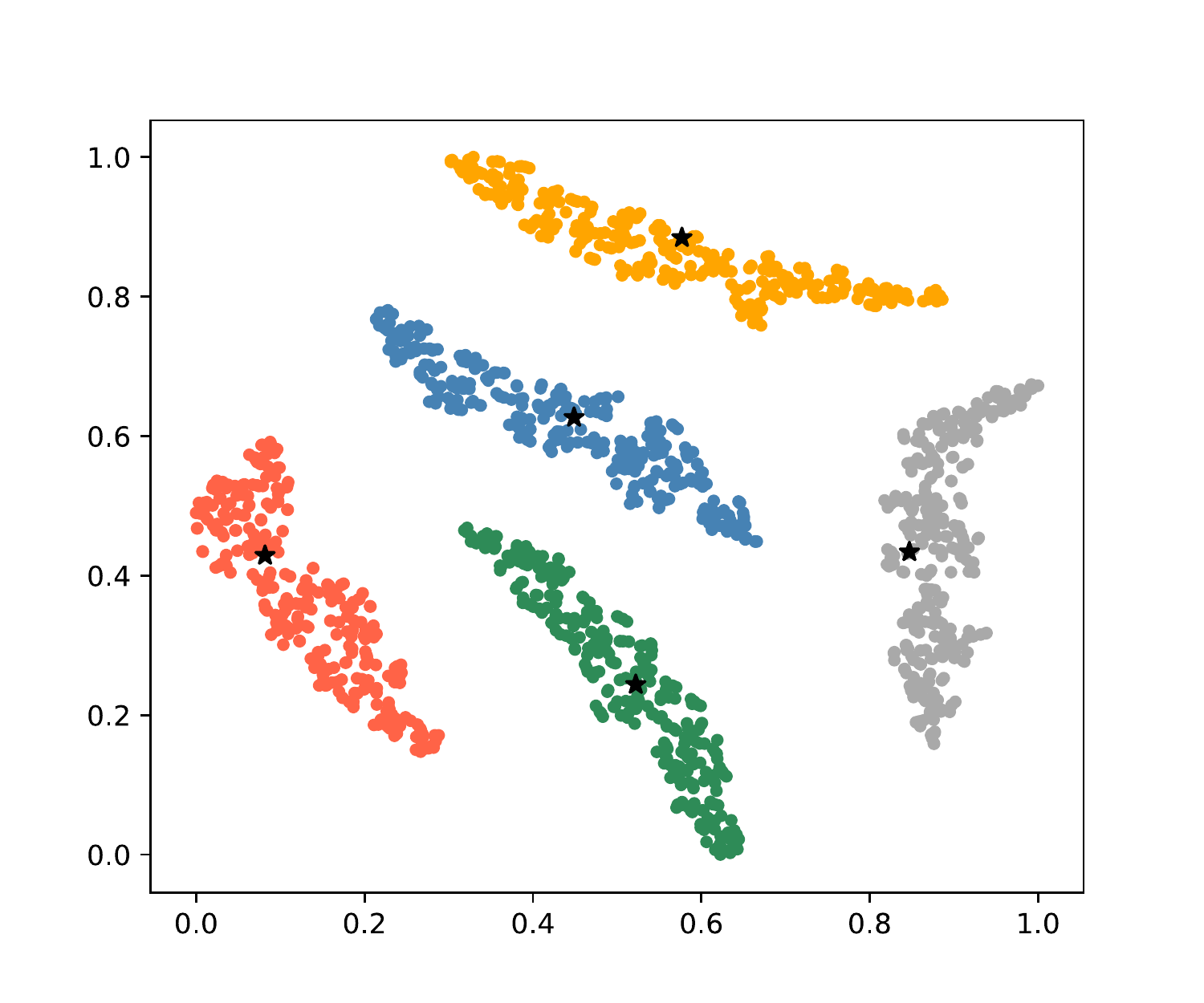}
  }
  \subfigure[DCVAE($\hat{x}$)(dis: 0.019)]{\label{ff}
    \includegraphics[scale=0.37]{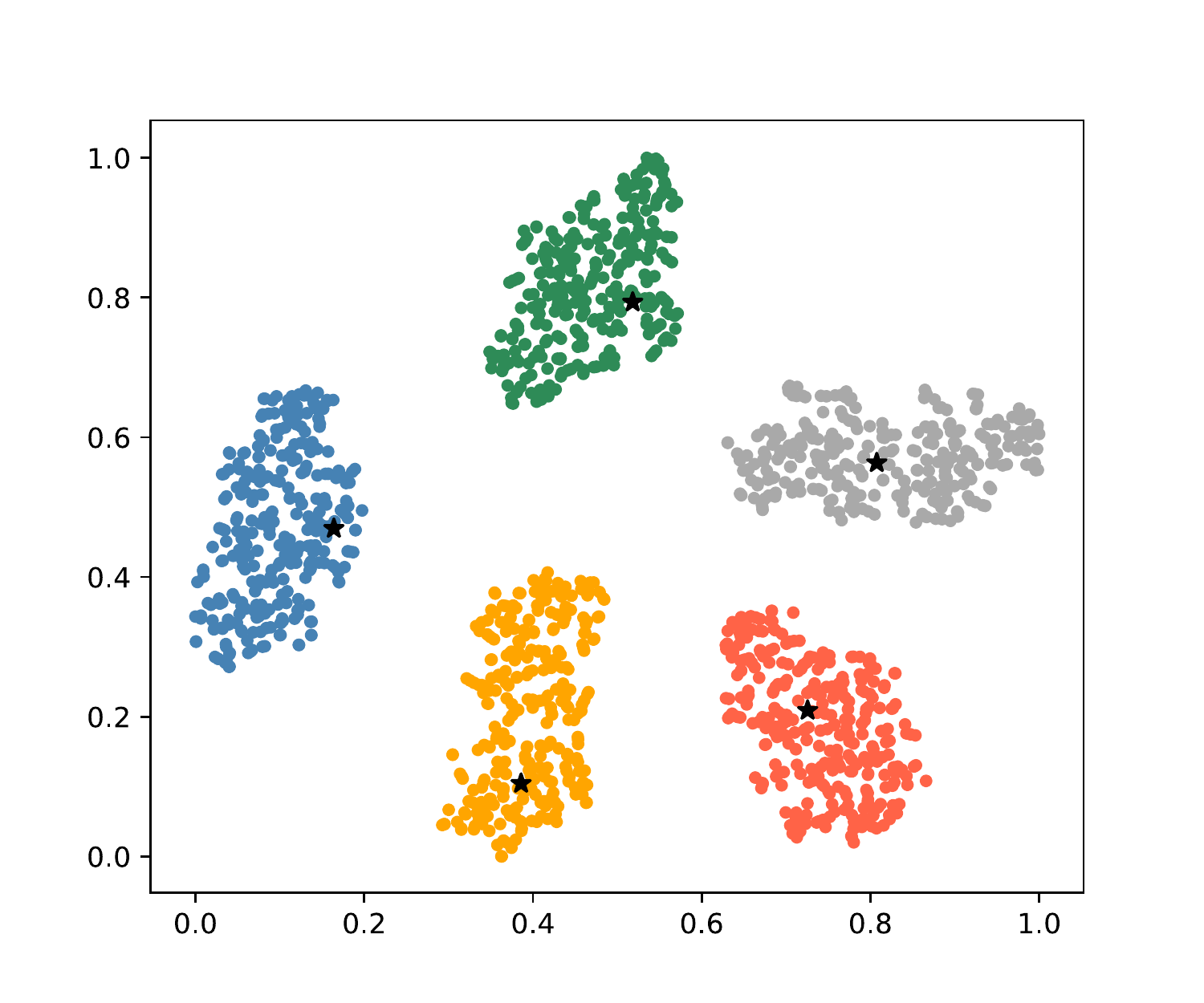}
  }
  \caption{The t-SNE visualization of synthetic features on miniImageNet. The class prototype is indicated by $\star$ in a 5-way 5-shot task. Figures~\ref{aa}-\ref{cc} visualize the distributions of data generated by the conventional Conditional Variational AutoEncoder (CVAE) via considering the single modality information as the feature generation condition, while Figures~\ref{dd}-\ref{ff} visualize the distributions of different modality features synthesized by our model. ``$s$;$v$'' represents the concatenation of semantic and visual information and ``dis'' indicates the mean of the Euclidean distances between the class prototype and the synthetic category prototype (the mean of the generated samples per category) in the few-shot classification task. Different colors are used for distinguishing the samples from different categories. The lower ``dis'' implies the smaller discrepancy between the distribution of real data and the distribution of synthesized data.}
  \label{tsne}
  \vspace{-0.5cm}
\end{figure*}

\subsubsection{Influences of Different Modules}
The proposed DCVAE consists of two essential modules, Dizygotic Feature Adaptive Mixture (DFAM) and Condition Cyclic Consistency (CCC). Thus, we perform an ablative study and empirically analyze the contributions of these two modules on the miniImageNet for 5-way setting. The analysis results are also shown in Table~\ref{as}. From observations, we can give the following conclusions: (1) the semantic representation plays a more important role for guiding the feature generation, particularly in one-shot scenario. (2) the DFAM module adaptively combines the features generated by two different conditions and achieves a performance gain of at least 1.31\%. (3) Combined with DFAM, CCC module further improves the performance from 65.38\% to 66.01\%. This indicates that condition cyclic consistency module can effectively ensure that the final synthetic features have sufficiently encoded both semantic and visual information, thus improving the quality of the generated features and few-shot classification performance.

\begin{table}[htb]
  \caption{Few-shot classification accuracy (in \%) on miniImageNet with different feature settings. ``$s$;$v$'' means the concatenation of semantic and visual information. The best results are in bold.}
  \centering
  \begin{tabular}{lcc}
    \toprule
    Methods                          & 5-way 1-shot   & 5-way 5-shot   \\
    \midrule
    CVAE($s$)                        & 63.71          & 76.52          \\
    CVAE($v$)                        & 56.42          & 73.35          \\
    CVAE($s$;$v$)                    & 61.10          & 74.15          \\
    \midrule
    DCVAE($x_{s}$)                   & 65.40          & 78.17          \\
    DCVAE($x_{v}$)                   & 62.32          & 78.03          \\
    DCVAE($\hat{x}$)                 & 66.01          & 78.35          \\
    DCVAE($x_{s} + x_{v}$)           & 65.73          & 78.26          \\
    DCVAE($x_{s} + \hat{x}$)         & \textbf{66.12} & \textbf{78.61} \\
    DCVAE($x_{v} + \hat{x}$)         & 65.84          & 78.29          \\
    DCVAE($x_{s} + x_{v} + \hat{x}$) & 66.06          & 78.54          \\
    \bottomrule
  \end{tabular}
  \label{fc}
  \vspace{-0.5cm}
\end{table}

\subsubsection{Effect of Different Feature Settings for Few-shot Classification Accuracy}
There are three kinds of synthetic features produced by our proposed model. They are visually conditional synthetic feature $x_v$, semantically conditional synthetic feature $x_s$ and the final hybrid synthetic feature $\hat{x}$ respectively. We conduct several experiments on miniImageNet via analyzing the influences of different combinations of these features to the performance of our model. Moreover, we build three extra CVAE models which only consider the semantic representation or the class prototype or the concatenation of these two representations as the unique condition for validating the effect of our proposed dizygotic symbiosis manner.

Table~\ref{fc} reports the results of the aforementioned experiments. From observations, here we can give the following conclusions:
(1) the dizygotic symbiosis manner boots the performance of the semantic and visual information-based feature generators both. For example, the $x_s$ and $x_v$ generated by DCVAE gain more 1.69\% and 5.90\% accuracies respectively than the ones generated by CVAE in 5-way 1-shot task.
(2) the performance of CVAE model with the concatenation of semantic and visual information as the condition is not significantly better than that of CVAE with the single condition. For example, the one-shot classification accuracy of such a simple model is 56.42\%~(visual only)$<$61.10\%$<$63.71\%~(semantic only). This means that if we have not elaborated a proper multi-modal learning fashion, the multi-modal version even cannot outperform the single one. The experimental results also imply that our method sufficiently exploit both of these two modalities of information, and present a considerable improvement in few-shot learning.
(3) the semantic representation plays a more important role for guiding the feature generation particularly in one-shot scenario, since the features generated based on semantics consistently achieve better results than the ones based on the visual information while their performance gap is shrunk along with the increase of the labeled samples. Such phenomenon again confirms the importance of semantics in few shot learning, which has been indicated in some previous works~\cite{xing2019adaptive}.
(4) the feature combinations always perform much better than the solo one, and $x_s+\hat{x}$ is the optimal combination. For example, $x_v+\hat{x}$ and $x_s+\hat{x}$ respectively improve 3.52\% and 0.72\% more accuracies.
(5) the final hybrid synthetic feature $\hat{x}$ is the best among the three generated features, however the performance gap between $x_s$ and $\hat{x}$ is very narrow.

\vspace{-0.3cm}
\subsection{Visualization of Synthetic Features}
We use the t-SNE~\cite{van2008visualizing} to visualize the data synthesized by Conditional Variational AutoEncoder (CVAE) and DCVAE with different modalities of prior knowledge in Figure~\ref{tsne}. The first row of the figure shows the distributions of data synthesized by CVAE with visual, semantic, combined modality prior knowledge from left to right. Similarly, the second row shows the ones of DCVAE with the same order. The $\star$ points represent the center of real samples (the category prototype) with respect to each category. In such a manner, the closer the distance between the center of the synthesized data and its corresponding point $\star$ is, the higher quality the synthesize features are. From observations, it is clear that the $\star$ points are always much closer to the center of data synthesized by DCVAE. For more explicitly showing such a fact, we report the mean of Euclidean distance between the center of synthesized data and its corresponding $\star$ over categories, which is referred as ``dis'' in Figure~\ref{tsne}. Clearly, such distances achieved by DCVAE are consistently shorter than the ones achieved by CVAE on all cases. In summary, all these visualization experimental results well validate the better feature generalization ability of DCVAE over CVAE.



\vspace{-0.1cm}
\section{Conclusion}
In this paper, we presented an ingenious multi-modal data augmentation method named Dizygotic Conditional Variational AutoEncoder (DCVAE) for few-shot learning. DCVAE pairs two Conditional Variational AutoEncoders (CVAEs) with a dizygotic symbiosis manner to utilize both the semantic and visual prior knowledge for conditioning the feature generation. It unifies the features generated based on different modality information to a final synthetic feature with an adaptive mixture mechanism, and employs a condition cyclic consistency module to keep the consistency between the original conditions and the conditions retrieved based on the final synthetic feature in both representation and function. DCVAE provides a new generative learning framework for data generation under multi-modal conditions. Extensive experimental results on three popular benchmarks demonstrate its superiority over the state-of-the-arts, and also validate that DCVAE is able to work well in various data modality configurations.


%



%
%

\ifCLASSOPTIONcaptionsoff
  \newpage
\fi



%


\bibliographystyle{IEEEtran}
\bibliography{reference}

\begin{thebibliography}{10}
\providecommand{\url}[1]{#1}
\csname url@samestyle\endcsname
\providecommand{\newblock}{\relax}
\providecommand{\bibinfo}[2]{#2}
\providecommand{\BIBentrySTDinterwordspacing}{\spaceskip=0pt\relax}
\providecommand{\BIBentryALTinterwordstretchfactor}{4}
\providecommand{\BIBentryALTinterwordspacing}{\spaceskip=\fontdimen2\font plus
\BIBentryALTinterwordstretchfactor\fontdimen3\font minus
  \fontdimen4\font\relax}
\providecommand{\BIBforeignlanguage}[2]{{%
\expandafter\ifx\csname l@#1\endcsname\relax
\typeout{** WARNING: IEEEtran.bst: No hyphenation pattern has been}%
\typeout{** loaded for the language `#1'. Using the pattern for}%
\typeout{** the default language instead.}%
\else
\language=\csname l@#1\endcsname
\fi
#2}}
\providecommand{\BIBdecl}{\relax}
\BIBdecl

\bibitem{longtail}
X.~Zhu, D.~Anguelov, and D.~Ramanan, ``Capturing long-tail distributions of
  object subcategories,'' in \emph{Proceedings of the IEEE Conference on
  Computer Vision and Pattern Recognition}, 2014, pp. 915--922.

\bibitem{xing2019adaptive}
C.~Xing, N.~Rostamzadeh, B.~Oreshkin, and P.~O. Pinheiro, ``Adaptive
  cross-modal few-shot learning,'' in \emph{Advances in Neural Information
  Processing Systems}, 2019, pp. 4847--4857.

\bibitem{pahde2021multimodal}
F.~Pahde, M.~Puscas, T.~Klein, and M.~Nabi, ``Multimodal prototypical networks
  for few-shot learning,'' in \emph{Proceedings of the IEEE/CVF Winter
  Conference on Applications of Computer Vision}, 2021, pp. 2644--2653.

\bibitem{snell2017prototypical}
J.~Snell, K.~Swersky, and R.~Zemel, ``Prototypical networks for few-shot
  learning,'' in \emph{Advances in neural information processing systems},
  2017, pp. 4077--4087.

\bibitem{xian2019f}
Y.~Xian, S.~Sharma, B.~Schiele, and Z.~Akata, ``f-vaegan-d2: A feature
  generating framework for any-shot learning,'' in \emph{Proceedings of the
  IEEE Conference on Computer Vision and Pattern Recognition}, 2019, pp.
  10\,275--10\,284.

\bibitem{vinyals2016matching}
O.~Vinyals, C.~Blundell, T.~Lillicrap, D.~Wierstra \emph{et~al.}, ``Matching
  networks for one shot learning,'' in \emph{Advances in neural information
  processing systems}, 2016, pp. 3630--3638.

\bibitem{khrulkov2020hyperbolic}
V.~Khrulkov, L.~Mirvakhabova, E.~Ustinova, I.~Oseledets, and V.~Lempitsky,
  ``Hyperbolic image embeddings,'' in \emph{Proceedings of the IEEE/CVF
  Conference on Computer Vision and Pattern Recognition}, 2020, pp. 6418--6428.

\bibitem{finn2017model}
C.~Finn, P.~Abbeel, and S.~Levine, ``Model-agnostic meta-learning for fast
  adaptation of deep networks,'' in \emph{International Conference on Machine
  Learning}.\hskip 1em plus 0.5em minus 0.4em\relax PMLR, 2017, pp. 1126--1135.

\bibitem{rusu2018meta}
A.~A. Rusu, D.~Rao, J.~Sygnowski, O.~Vinyals, R.~Pascanu, S.~Osindero, and
  R.~Hadsell, ``Meta-learning with latent embedding optimization,'' \emph{arXiv
  preprint arXiv:1807.05960}, 2018.

\bibitem{simon2020adaptive}
C.~Simon, P.~Koniusz, R.~Nock, and M.~Harandi, ``Adaptive subspaces for
  few-shot learning,'' in \emph{Proceedings of the IEEE/CVF Conference on
  Computer Vision and Pattern Recognition}, 2020, pp. 4136--4145.

\bibitem{zhang2018metagan}
R.~Zhang, T.~Che, Z.~Ghahramani, Y.~Bengio, and Y.~Song, ``Metagan: An
  adversarial approach to few-shot learning,'' in \emph{Advances in Neural
  Information Processing Systems}, 2018, pp. 2365--2374.

\bibitem{zhang2019few}
H.~Zhang, J.~Zhang, and P.~Koniusz, ``Few-shot learning via saliency-guided
  hallucination of samples,'' in \emph{Proceedings of the IEEE Conference on
  Computer Vision and Pattern Recognition}, 2019, pp. 2770--2779.

\bibitem{li2020adversarial}
K.~Li, Y.~Zhang, K.~Li, and Y.~Fu, ``Adversarial feature hallucination networks
  for few-shot learning,'' in \emph{Proceedings of the IEEE/CVF Conference on
  Computer Vision and Pattern Recognition}, 2020, pp. 13\,470--13\,479.

\bibitem{chen2020diversity}
M.~Chen, Y.~Fang, X.~Wang, H.~Luo, Y.~Geng, X.~Zhang, C.~Huang, W.~Liu, and
  B.~Wang, ``Diversity transfer network for few-shot learning,'' in
  \emph{Proceedings of the AAAI Conference on Artificial Intelligence},
  vol.~34, no.~07, 2020, pp. 10\,559--10\,566.

\bibitem{Zitian}
Z.~Chen, Y.~Fu, Y.-X. Wang, L.~Ma, W.~Liu, and M.~Hebert, ``Image deformation
  meta-networks for one-shot learning,'' in \emph{Proceedings of the IEEE
  Conference on Computer Vision and Pattern Recognition}, 2019, pp. 8680--8689.

\bibitem{chen2019image}
Z.~Chen, Y.~Fu, K.~Chen, and Y.-G. Jiang, ``Image block augmentation for
  one-shot learning,'' in \emph{Proceedings of the AAAI Conference on
  Artificial Intelligence}, vol.~33, 2019, pp. 3379--3386.

\bibitem{xian2017zero}
Y.~Xian, B.~Schiele, and Z.~Akata, ``Zero-shot learning-the good, the bad and
  the ugly,'' in \emph{Proceedings of the IEEE Conference on Computer Vision
  and Pattern Recognition}, 2017, pp. 4582--4591.

\bibitem{avae}
E.~Schonfeld, S.~Ebrahimi, S.~Sinha, T.~Darrell, and Z.~Akata, ``Generalized
  zero-and few-shot learning via aligned variational autoencoders,'' in
  \emph{Proceedings of the IEEE Conference on Computer Vision and Pattern
  Recognition}, 2019, pp. 8247--8255.

\bibitem{schwartz2019baby}
E.~Schwartz, L.~Karlinsky, R.~Feris, R.~Giryes, and A.~M. Bronstein, ``Baby
  steps towards few-shot learning with multiple semantics,'' \emph{arXiv
  preprint arXiv:1906.01905}, 2019.

\bibitem{tsai2018learning}
Y.-H.~H. Tsai, P.~P. Liang, A.~Zadeh, L.-P. Morency, and R.~Salakhutdinov,
  ``Learning factorized multimodal representations,'' in \emph{International
  Conference on Learning Representations}, 2019.

\bibitem{shi2021relating}
Y.~Shi, B.~Paige, P.~Torr, and S.~N, ``Relating by contrasting: A
  data-efficient framework for multimodal generative models,'' in
  \emph{International Conference on Learning Representations}, 2021.

\bibitem{ma2021smil}
M.~Ma, J.~Ren, L.~Zhao, S.~Tulyakov, C.~Wu, and X.~Peng, ``Smil: Multimodal
  learning with severely missing modality,'' in \emph{Proceedings of the AAAI
  Conference on Artificial Intelligence}, vol.~35, no.~3, 2021, pp. 2302--2310.

\bibitem{qi2018low}
H.~Qi, M.~Brown, and D.~G. Lowe, ``Low-shot learning with imprinted weights,''
  in \emph{Proceedings of the IEEE conference on computer vision and pattern
  recognition}, 2018, pp. 5822--5830.

\bibitem{tadam}
B.~Oreshkin, P.~R. L{\'o}pez, and A.~Lacoste, ``Tadam: Task dependent adaptive
  metric for improved few-shot learning,'' in \emph{Advances in Neural
  Information Processing Systems}, 2018, pp. 721--731.

\bibitem{sung2018learning}
F.~Sung, Y.~Yang, L.~Zhang, T.~Xiang, P.~H. Torr, and T.~M. Hospedales,
  ``Learning to compare: Relation network for few-shot learning,'' in
  \emph{Proceedings of the IEEE Conference on Computer Vision and Pattern
  Recognition}, 2018, pp. 1199--1208.

\bibitem{abdelaziz2021few}
M.~Abdelaziz and Z.~Zhang, ``Few-shot learning with saliency maps as additional
  visual information,'' \emph{Multimedia Tools and Applications}, vol.~80,
  no.~7, pp. 10\,491--10\,508, 2021.

\bibitem{ren2019incremental}
M.~Ren, R.~Liao, E.~Fetaya, and R.~Zemel, ``Incremental few-shot learning with
  attention attractor networks,'' in \emph{Advances in Neural Information
  Processing Systems}, 2019, pp. 5275--5285.

\bibitem{sun2019meta}
Q.~Sun, Y.~Liu, T.-S. Chua, and B.~Schiele, ``Meta-transfer learning for
  few-shot learning,'' in \emph{Proceedings of the IEEE conference on computer
  vision and pattern recognition}, 2019, pp. 403--412.

\bibitem{oh2021boil}
J.~Oh, H.~Yoo, C.~Kim, and S.-Y. Yun, ``Boil: Towards representation change for
  few-shot learning,'' in \emph{International Conference on Learning
  Representations}, 2021.

\bibitem{wang2018low}
Y.-X. Wang, R.~Girshick, M.~Hebert, and B.~Hariharan, ``Low-shot learning from
  imaginary data,'' in \emph{Proceedings of the IEEE conference on computer
  vision and pattern recognition}, 2018, pp. 7278--7286.

\bibitem{luo2021few}
Q.~Luo, L.~Wang, J.~Lv, S.~Xiang, and C.~Pan, ``Few-shot learning via feature
  hallucination with variational inference,'' in \emph{Proceedings of the
  IEEE/CVF Winter Conference on Applications of Computer Vision}, 2021, pp.
  3963--3972.

\bibitem{chen2019multi}
Z.~Chen, Y.~Fu, Y.~Zhang, Y.-G. Jiang, X.~Xue, and L.~Sigal, ``Multi-level
  semantic feature augmentation for one-shot learning,'' \emph{IEEE
  Transactions on Image Processing}, vol.~28, no.~9, pp. 4594--4605, 2019.

\bibitem{yoon2019tapnet}
S.~W. Yoon, J.~Seo, and J.~Moon, ``Tapnet: Neural network augmented with
  task-adaptive projection for few-shot learning,'' in \emph{International
  Conference on Machine Learning}, 2019, pp. 7115--7123.

\bibitem{razavi2019generating}
A.~Razavi, A.~van~den Oord, and O.~Vinyals, ``Generating diverse high-fidelity
  images with vq-vae-2,'' in \emph{Advances in Neural Information Processing
  Systems}, 2019, pp. 14\,866--14\,876.

\bibitem{zhao2018unsupervised}
T.~Zhao, K.~Lee, and M.~Eskenazi, ``Unsupervised discrete sentence
  representation learning for interpretable neural dialog generation,'' in
  \emph{Proceedings of the 56th Annual Meeting of the Association for
  Computational Linguistics (Volume 1: Long Papers)}, 2018, pp. 1098--1107.

\bibitem{makhzani2015adversarial}
A.~Makhzani, J.~Shlens, N.~Jaitly, I.~Goodfellow, and B.~Frey, ``Adversarial
  autoencoders,'' \emph{arXiv preprint arXiv:1511.05644}, 2015.

\bibitem{verma2018generalized}
V.~K. Verma, G.~Arora, A.~Mishra, and P.~Rai, ``Generalized zero-shot learning
  via synthesized examples,'' in \emph{Proceedings of the IEEE conference on
  computer vision and pattern recognition}, 2018, pp. 4281--4289.

\bibitem{hao2019collect}
F.~Hao, F.~He, J.~Cheng, L.~Wang, J.~Cao, and D.~Tao, ``Collect and select:
  Semantic alignment metric learning for few-shot learning,'' in
  \emph{Proceedings of the IEEE International Conference on Computer Vision},
  2019, pp. 8460--8469.

\bibitem{chen2019closer}
W.-Y. Chen, Y.-C. Liu, Z.~Kira, Y.-C.~F. Wang, and J.-B. Huang, ``A closer look
  at few-shot classification,'' \emph{arXiv preprint arXiv:1904.04232}, 2019.

\bibitem{chen2021hierarchical}
C.~Chen, K.~Li, W.~Wei, J.~T. Zhou, and Z.~Zeng, ``Hierarchical graph neural
  networks for few-shot learning,'' \emph{IEEE Transactions on Circuits and
  Systems for Video Technology}, 2021.

\bibitem{huang2021local}
H.~Huang, Z.~Wu, W.~Li, J.~Huo, and Y.~Gao, ``Local descriptor-based
  multi-prototype network for few-shot learning,'' \emph{Pattern Recognition},
  vol. 116, p. 107935, 2021.

\bibitem{lazarou2021few}
M.~Lazarou, Y.~Avrithis, and T.~Stathaki, ``Few-shot learning via tensor
  hallucination,'' \emph{International Conference on Learning Representations
  Workshop}, 2021.

\bibitem{schwartz2018delta}
E.~Schwartz, L.~Karlinsky, J.~Shtok, S.~Harary, M.~Marder, A.~Kumar, R.~Feris,
  R.~Giryes, and A.~Bronstein, ``Delta-encoder: an effective sample synthesis
  method for few-shot object recognition,'' in \emph{Advances in Neural
  Information Processing Systems}, 2018, pp. 2845--2855.

\bibitem{lee2020self}
H.~Lee, S.~J. Hwang, and J.~Shin, ``Self-supervised label augmentation via
  input transformations,'' in \emph{International Conference on Machine
  Learning}.\hskip 1em plus 0.5em minus 0.4em\relax PMLR, 2020, pp. 5714--5724.

\bibitem{guo2020attentive}
Y.~Guo and N.-M. Cheung, ``Attentive weights generation for few shot learning
  via information maximization,'' in \emph{Proceedings of the IEEE/CVF
  Conference on Computer Vision and Pattern Recognition}, 2020, pp.
  13\,499--13\,508.

\bibitem{li2020asymmetric}
W.~Li, L.~Wang, J.~Huo, Y.~Shi, Y.~Gao, and J.~Luo, ``Asymmetric distribution
  measure for few-shot learning,'' \emph{Proceedings of the Twenty-Ninth
  International Joint Conference on Artificial Intelligence}, 2020.

\bibitem{majumder2021revisiting}
O.~Majumder, A.~Ravichandran, S.~Maji, M.~Polito, R.~Bhotika, and S.~Soatto,
  ``Revisiting contrastive learning for few-shot classification,'' \emph{arXiv
  preprint arXiv:2101.11058}, 2021.

\bibitem{xu2021attentional}
W.~Xu, yifan xu, H.~Wang, and Z.~Tu, ``Attentional constellation nets for
  few-shot learning,'' in \emph{International Conference on Learning
  Representations}, 2021.

\bibitem{li2019few}
A.~Li, T.~Luo, T.~Xiang, W.~Huang, and L.~Wang, ``Few-shot learning with global
  class representations,'' in \emph{Proceedings of the IEEE International
  Conference on Computer Vision}, 2019, pp. 9715--9724.

\bibitem{das2019two}
D.~Das and C.~G. Lee, ``A two-stage approach to few-shot learning for image
  recognition,'' \emph{IEEE Transactions on Image Processing}, vol.~29, pp.
  3336--3350, 2019.

\bibitem{ravichandran2019few}
A.~Ravichandran, R.~Bhotika, and S.~Soatto, ``Few-shot learning with embedded
  class models and shot-free meta training,'' in \emph{Proceedings of the
  IEEE/CVF International Conference on Computer Vision}, 2019, pp. 331--339.

\bibitem{lee2019meta}
K.~Lee, S.~Maji, A.~Ravichandran, and S.~Soatto, ``Meta-learning with
  differentiable convex optimization,'' in \emph{Proceedings of the IEEE/CVF
  Conference on Computer Vision and Pattern Recognition}, 2019, pp.
  10\,657--10\,665.

\bibitem{kim2020model}
J.~Kim, H.~Kim, and G.~Kim, ``Model-agnostic boundary-adversarial sampling for
  test-time generalization in few-shot learning,'' in \emph{European Conference
  on Computer Vision (ECCV 2020). Online}, 2020, pp. 5025--2.

\bibitem{wang2020instance}
Y.~Wang, C.~Xu, C.~Liu, L.~Zhang, and Y.~Fu, ``Instance credibility inference
  for few-shot learning,'' in \emph{Proceedings of the IEEE/CVF Conference on
  Computer Vision and Pattern Recognition}, 2020, pp. 12\,836--12\,845.

\bibitem{tian2020rethinking}
Y.~Tian, Y.~Wang, D.~Krishnan, J.~B. Tenenbaum, and P.~Isola, ``Rethinking
  few-shot image classification: a good embedding is all you need?''
  \emph{European Conference on Computer Vision}, 2020.

\bibitem{russakovsky2015imagenet}
O.~Russakovsky, J.~Deng, H.~Su, J.~Krause, S.~Satheesh, S.~Ma, Z.~Huang,
  A.~Karpathy, A.~Khosla, M.~Bernstein \emph{et~al.}, ``Imagenet large scale
  visual recognition challenge,'' \emph{International journal of computer
  vision}, vol. 115, no.~3, pp. 211--252, 2015.

\bibitem{wah2011caltech}
C.~Wah, S.~Branson, P.~Welinder, P.~Perona, and S.~Belongie, ``The caltech-ucsd
  birds-200-2011 dataset,'' 2011.

\bibitem{krizhevsky2009learning}
A.~Krizhevsky, G.~Hinton \emph{et~al.}, ``Learning multiple layers of features
  from tiny images,'' 2009.

\bibitem{pennington2014glove}
J.~Pennington, R.~Socher, and C.~D. Manning, ``Glove: Global vectors for word
  representation,'' in \emph{Proceedings of the 2014 conference on empirical
  methods in natural language processing (EMNLP)}, 2014, pp. 1532--1543.

\bibitem{van2008visualizing}
L.~Van~der Maaten and G.~Hinton, ``Visualizing data using t-sne.''
  \emph{Journal of machine learning research}, vol.~9, no.~11, 2008.

\end{thebibliography}

\end{document}